%% file: 0-main.tex
\RequirePackage{fix-cm}
\documentclass[smallextended]{svjour3}       
\smartqed  
\usepackage{graphicx}

\usepackage[commentmarkup=todo]{changes}

\usepackage{latexsym}
\usepackage{times}
\usepackage{xurl}
\usepackage{booktabs}
\usepackage{makecell}
\usepackage{tabularx}
\usepackage{array}
\newcolumntype{C}{>{\centering\arraybackslash}p{4.7em}}
\usepackage{multirow}
\usepackage{colortbl}
\usepackage{amsmath}
\usepackage{amsfonts}
\usepackage{mathabx}
\usepackage{subcaption}
\usepackage{url}
\usepackage{caption}
\usepackage{arydshln}
\usepackage{comment}
\usepackage{todonotes}

\RequirePackage[bookmarks, bookmarksopen=true, plainpages=false, pdfpagelabels, pdfpagelayout=SinglePage, breaklinks = true]{hyperref}

\DeclareMathOperator*{\argmin}{arg\,min}
\PassOptionsToPackage{hyphens}{url}

\renewcommand{\deleted}[2][]{} 

\renewcommand{\added}[2][]{#2}

\begin{document}

\title{On Cross-Lingual Retrieval with Multilingual Text Encoders
}


\author{Robert Litschko \and
Ivan Vuli\'{c} \and
Simone Paolo Ponzetto \and
Goran Glava\v{s}}



\institute{Robert Litschko, Simone Paolo Ponzetto and Goran Glava\v{s} \at
           University of Mannheim \\
           \email{\{litschko,simone,goran\}@informatik.uni-mannheim.de}           
           \and
           Ivan Vuli\'{c} \at
           Language Technology Lab, University of Cambridge \\
           \email{iv250@cam.ac.uk}           %
}

\date{Received: date / Accepted: date}

\maketitle

\begin{abstract}
\deleted[comment=abstract too long]{Pretrained multilingual text encoders based on neural \textit{transformer architectures}, such as multilingual BERT (mBERT) and XLM, have recently displayed unparalleled performance on a myriad of language understanding tasks, when fine-tuned in a supervised fashion on annotated task-specific data. As a consequence, they have \textit{de facto} become a default paradigm for multilingual representation learning and cross-lingual transfer of natural language processing models, rendering cross-lingual word embedding spaces (CLWEs) effectively obsolete.}
\deleted{However, questions remain to which extent these findings generalize to ad-hoc cross-lingual information retrieval (CLIR) settings that are most frequently met in practice. There, one commonly operates with a limited amount of relevance judgments (i.e., labeled data), which in most cases originate from a (somewhat) different domain.}
\added{Pretrained multilingual text encoders based on neural \textit{transformer architectures}, such as multilingual BERT (mBERT) and XLM, have recently become a default paradigm for cross-lingual transfer of natural language processing models, rendering cross-lingual word embedding spaces (CLWEs) effectively obsolete.}
\deleted{Therefore, in} \added{In} this work we present a systematic empirical study focused on the suitability of the state-of-the-art multilingual encoders for cross-lingual document and sentence retrieval tasks across a number of diverse language pairs. We first treat these models as multilingual text encoders and benchmark their performance in unsupervised ad-hoc sentence- and document-level CLIR. In contrast to supervised language understanding, our results indicate that for unsupervised document-level CLIR -- a setup with no relevance judgments for IR-specific fine-tuning -- pretrained multilingual encoders on average fail to significantly outperform earlier models based on CLWEs. For sentence-level retrieval, we do obtain state-of-the-art performance: the peak scores, however, are met by multilingual encoders that have been further specialized, in a supervised fashion, for sentence understanding tasks, rather than using their vanilla `off-the-shelf' variants. Following these results, we introduce localized relevance matching for document-level CLIR, \added{where we independently score a query against document sections.} \deleted{compare document segments and sentences independently against the query and then max-pool their relevance scores.}  
In the second part, we evaluate multilingual encoders fine-tuned in a supervised fashion (i.e., we \textit{learn to rank}) on English relevance data in a series of zero-shot language and domain transfer CLIR experiments. Our results show that, despite the supervision, and due to the domain and language shift, supervised re-ranking rarely improves the performance of multilingual transformers as unsupervised base rankers. Finally, only with in-domain contrastive fine-tuning (i.e., same domain, only language transfer), we manage to improve the ranking quality. We uncover substantial empirical differences between cross-lingual retrieval results and results of (zero-shot) cross-lingual transfer for monolingual retrieval in target languages, which point to ``monolingual overfitting'' of retrieval models trained on monolingual (English) data, even if they are based on multilingual transformers. 

\end{abstract}

\input{1-introduction}
\input{2-related_work}
\input{3-models}
\input{4-unsup}

\input{5-l2r}
\input{6-conclusion}

\begin{acknowledgements}
The work of Ivan Vuli\'{c} is supported by the ERC PoC Grant MultiConvAI (no. 957356). Goran Glava\v{s} is supported by the Baden-Württemberg Ministry of Economic Affairs, Labour and Tourism through the Multi\textsuperscript{2}ConvAI grant (AI-Innovation Programme). The authors are grateful to Sean MacAvaney for his help with evaluating the zero-shot mBERT ranker and to Nils Reimers for his feedback on sentence encoders. 
\end{acknowledgements}

%
%


\bibliographystyle{spmpsci}
\bibliography{references}

%
%

\end{document}

%% file: 1-introduction.tex
\section{Introduction}

Cross-lingual information retrieval (CLIR) systems respond to queries in a source language by retrieving relevant documents in another, target language. Their success is typically hindered by data scarcity: they operate in challenging low-resource settings without sufficient labeled training data, i.e., human relevance judgments, to build reliable in-domain supervised models (e.g., neural matching models for pairwise retrieval \cite{10.1145/3397271.3401322,jiang2020cross}). This motivates the need for robust, resource-lean CLIR approaches: (1) unsupervised CLIR models and/or (2) transfer of supervised rankers across domains and languages, i.e., from resource-rich to resource-lean setups.

In previous work, Litschko et al.\ \cite{litschko2019evaluating} have shown that language transfer by means of cross-lingual embedding spaces (CLWEs) can be used to yield state-of-the-art performance in a range of unsupervised ad-hoc CLIR setups. This approach uses very weak cross-lingual (in this case, bilingual) supervision (i.e., only a bilingual dictionary spanning 1K-5K word translation pairs), or even no bilingual supervision at all, in order to learn a mapping that aligns two monolingual word embedding spaces \cite{glavas-etal-2019-properly,Vulic:2019emnlp}. Put simply, this enables casting CLIR tasks as `monolingual tasks in the shared (CLWE) space': at retrieval time both queries and documents are represented as simple aggregates of their constituent CLWEs. However, this approach, by limitations of static CLWEs, cannot capture and handle polysemy in the underlying text representations, and captures only ``static'' word-level semantics. \textit{Contextual text representation models} alleviate this issue \cite{liu2020survey} because they encode occurrences of the same word differently depending on its context.   


Such contextual dynamic representations are obtained via deep neural models pretrained on large text collections through general objectives such as (masked) language modeling \cite{devlin-etal-2019-bert,Liu:2019roberta}. Multilingual text encoders pretrained on 100+ languages, such as multilingual BERT (mBERT) \cite{devlin-etal-2019-bert} or XLM(-R) \cite{xlm,conneau2020unsupervised}, have become a \textit{de facto} standard for multilingual representation learning and cross-lingual transfer in natural language processing (NLP). These models demonstrate state-of-the-art performance in a wide range of supervised language understanding and language generation tasks \cite{xcopa,xglue}: the general-purpose language knowledge obtained during pretraining is successfully specialized using task-specific training (i.e., fine-tuning). Multilingual transformers have been rendered especially effective in zero-shot transfer settings: a typical \textit{modus operandi} is fine-tuning a pretrained multilingual encoder with task-specific data of a source language (typically English) and then using it directly in a target language. The effectiveness of cross-lingual transfer with multilingual transformers, however, has more recently been shown to highly depend on the typological proximity between languages as well as the size of the pretraining corpora in the target language \cite{hu2020xtreme,lauscher2020zero,zhao2021few}.     

It is unclear, however, whether these general-purpose multilingual text encoders can be used directly for ad-hoc CLIR without any additional supervision (i.e., cross-lingual relevance judgments). Further, can they outperform unsupervised CLIR approaches based on static CLWEs \cite{litschko2019evaluating}? How do they perform depending on the (properties of the) language pair at hand? How can we encode useful semantic information using these models, and do different ``encoding variants'' (see later \S\ref{s:mencoders}) yield different retrieval results? Are there performance differences in unsupervised sentence-level versus document-level CLIR tasks? Can we boost performance by relying on sentence encoders that are specialized towards dealing with sentence-level understanding in particular? Finally, can we improve ad-hoc CLIR in our target setups by fine-tuning multilingual encoders on relevance judgments from different document collections (i.e., domains) and languages (e.g., by exploiting available monolingual English relevance judgments from other collections)? 

In order to address all these questions, we present a systematic empirical study and profile the suitability of state-of-the-art pretrained multilingual encoders for different CLIR tasks and diverse language pairs, across unsupervised, supervised, and transfer setups. We evaluate state-of-the-art general-purpose pretrained multilingual encoders (mBERT \cite{devlin-etal-2019-bert} and XLM \cite{xlm}) with a range of encoding variants, and also compare them to provenly robust CLIR approaches based on static CLWEs, as well as to specialized variants of multilingual encoders fine-tuned to encode sentence semantics \cite[\textit{inter alia}]{artetxe2019LASER,feng2020language-labse,reimers2020making-distil}. Finally, we compare the unsupervised CLIR approaches based on these multilingual transformers with their counterparts fine-tuned on English relevance signal from different domains/collections. Our key contributions and findings are summarized as follows:

%
%
%

\vspace{1.2mm}
\noindent \textbf{(1)} We empirically validate (\S\ref{sec:base_results_docs}) that, without any task-specific fine-tuning, multilingual encoders such as mBERT and XLM fail to outperform CLIR approaches based on static CLWEs. Their performance also crucially depends on how one encodes semantic information with the models (e.g., treating them as sentence/document encoders directly versus averaging over constituent words and/or subwords). 

\vspace{1.2mm}
\noindent \textbf{(2)} We show that multilingual sentence encoders, fine-tuned on labeled data from sentence pair tasks like natural language inference or semantic text similarity as well as using parallel sentences, substantially outperform general-purpose models (mBERT and XLM) in sentence-level CLIR (\S\ref{sec:sentence-level-clir}); further, they can be leveraged for localized relevance matching and in such a pooling setup improve the performance of unsupervised document-level CLIR (\S\ref{sec:localrelmatching}).


\vspace{1.2mm}
\noindent \textbf{(3)} Supervised neural rankers (also based on multilingual transformers like mBERT) trained on English relevance judgments from different collections (i.e., zero-shot language and domain transfer) do not surpass the best-performing unsupervised CLIR approach based on multilingual sentence encoders, either as standalone rankers or as re-rankers of the initial ranking produced by the unsupervised CLIR model based on multilingual sentence encoders (\S\ref{sec:reranking}).    

\vspace{1.2mm}

\noindent \textbf{(4)} In-domain fine-tuning of the best-performing unsupervised transformer \cite{reimers2020making-distil} (i.e., zero-shot language transfer, no domain transfer) -- yields considerable gains over the original unsupervised ranker (\S\ref{sec:finetuningdistilmbert}). This renders fine-tuning with little in-domain data more beneficial than transferring models trained on large-scale out-of-domain datasets. 


\vspace{1.2mm}

\noindent \textbf{(5)} Finally, we show that fine-tuning supervised CLIR models based on multilingual transformers on monolingual (English) data leads to a type of ``overfitting'' to monolingual retrieval (\S\ref{sec:multi_vs_cross}): \deleted{such models transfer much better to monolingual retrieval tasks in (unseen) target languages than to cross-lingual retrieval tasks} \added{We empirically show that language transfer in IR is more difficult in true cross-lingual IR settings, in which query and documents are in different languages, as opposed to monolingual IR in a different (target) language.}

\vspace{1.2mm}

\added[comment=Moved from footnote and rephrased slightly]{This manuscript is an extension of the article \textit{``Evaluating Multilingual Text Encoders for Unsupervised Cross-Lingual Retrieval''} published in the Proceedings of the 43rd European Conference on Information Retrieval (ECIR) \cite{ecir2021-litschko}, where we evaluated multilingual encoders exclusively in unsupervised CLIR. In this work we, first and foremost, extend the scope of the work to supervised IR settings, and investigate how (English, in-domain or out-of-domain) relevance annotations can be leveraged to fine-tune supervised rankers based on multilingual text encoders (e.g., multilingual BERT). To this end, we evaluate document-level CLIR performance of (1) two standard pointwise learning-to-rank (L2R) models based on multilingual BERT and trained on large-scale English corpora and (2) a multilingual encoder fine-tuned via contrastive metric-based learning on small in-domain relevance dataset; we demonstrate that only the latter offers consistent performance gains over unsupervised CLIR with the same multilingual encoders. Pointwise L2R and contrastive fine-tuning models are described in Section 3.4. Section 5 provides detailed experimental evaluation of those models on several document-level CLIR tasks.}

We believe that this extensive empirical study offers plenty of valuable new insights for researchers and practitioners who work in the challenging landscape of cross-lingual information retrieval tasks.

%
%
%
%


%% file: 2-related_work.tex
\section{Related Work}
\label{sec:rw}

\textbf{Self-Supervised Pretraining and Transfer Learning.} Recently, research on universal sentence representations and transfer learning has gained much traction. InferSent \cite{conneau-EtAl:2017:EMNLP2017} transfers the encoder of a model trained on natural language inference to other tasks, while USE \cite{cer2018universal} extends this idea to a multi-task learning setting. More recent work explores self-supervised neural Transformer-based \cite{vaswani2017attention} models, all based on (causal or masked) language modeling (LM) objectives, such as BERT \cite{devlin-etal-2019-bert}, RoBERTa \cite{Liu:2019roberta}, GPT \cite{radford2019language,brown2020language}, and XLM \cite{xlm}.\footnote{Note that self-supervised learning can come in different flavors depending on the training objective \cite{Clark:2020iclr}, but language modeling objectives still seem to be the most popular choice.} Results on benchmarks such as GLUE \cite{wang2018glue} and SentEval \cite{conneau2018senteval} indicate that these models can yield impressive (sometimes human-level) performance in supervised Natural Language Understanding (NLU) and Generation (NLG) tasks. These models have become \emph{de facto} standard and omnipresent text representation models in NLP. In supervised monolingual IR, self-supervised LMs have been employed as contextualized word encoders \cite{macavaney2019cedr}, or fine-tuned as pointwise and pairwise rankers \cite{nogueira2019multi}.
%
%

\vspace{1.8mm}
\noindent
\textbf{Multilingual Text Encoders} based on the (masked) LM objectives have also been massively adopted in multilingual and cross-lingual NLP and IR applications. A multilingual extension of BERT (mBERT) is trained with a shared subword vocabulary on a single multilingual corpus obtained as concatenation of large monolingual data in 104 languages. The XLM model \cite{xlm} extends this idea and proposes natively cross-lingual LM pretraining, combining causal language modeling (CLM) and translation language modeling (TLM).\footnote{In CLM, the model is trained to predict the probability of a word given the previous words in a sentence. TLM is a cross-lingual variant of standard masked LM (MLM), with the core difference that the model is given pairs of parallel sentences and allowed to attend to the aligned sentence when reconstructing a word in the current sentence.}  Strong performance of these models in supervised settings is confirmed across a range of tasks on multilingual benchmarks such as XGLUE \cite{xglue} and XTREME \cite{hu2020xtreme}. However, recent work \cite{reimers2020making-distil,cao2019multilingual} has indicated that these general-purpose models do not yield strong results when used as out-of-the-box text encoders in an unsupervised transfer learning setup. We further investigate these preliminaries, and confirm this finding also for unsupervised ad-hoc CLIR tasks. 

Multilingual text encoders have already found applications in document-level CLIR. Jiang et al. \cite{jiang2020cross} use mBERT as a matching model by feeding pairs of English queries and foreign language documents. MacAvaney et al. \cite{macavaney2020teaching} use mBERT in a zero-shot setting, where they train a retrieval model on top of mBERT on English relevance data and apply it on a different language.



\vspace{1.8mm}
\noindent
\textbf{Specialized Multilingual Sentence Encoders.}
An extensive body of work focuses on inducing multilingual encoders that capture sentence meaning. In \cite{artetxe2019LASER}, the multilingual encoder of a sequence-to-sequence model is shared across languages and optimized to be language-agnostic, whereas Guo et al. \cite{guo-etal-2018-effective} rely on a dual Transformer-based encoder architecture instead (with tied/shared parameters) to represent parallel sentences. Rather than optimizing for translation performance directly, their approach minimizes the cosine distance between parallel sentences. A ranking softmax loss is used to classify the correct (i.e., aligned) sentence in the other language from negative samples (i.e., non-aligned sentences). In \cite{yang2019improving}, this approach is extended by using a bidirectional dual encoder and adding an additive margin softmax function, which serves to push away non-translation-pairs in the shared embedding space. The dual-encoder approach is now widely adopted  \cite{guo-etal-2018-effective,yang-etal-2020-multilingual-muse,feng2020language-labse,reimers2020making-distil,zhao2020inducing}, and yields state-of-the-art multilingual sentence encoders which excel in sentence-level NLU tasks.

Other recent approaches propose input space normalization, and using parallel data to re-align mBERT and XLM \cite{zhao2020inducing,cao2019multilingual}, or using a teacher-student framework where a student model is trained to imitate the output of the teacher network while preserving high similarity of translation pairs \cite{reimers2020making-distil}. In \cite{yang-etal-2020-multilingual-muse}, authors combine multi-task learning with a translation bridging task to train a universal sentence encoder. We benchmark a series of representative sentence encoders in this article; their brief descriptions are provided in \S\ref{sec:sent-specialized-models}.


\vspace{1.8mm}
\noindent
\textbf{Neural Learning-to-Rank.} In the context of neural retrieval the vast majority of rankers can be broadly classified into the two paradigms of (i) Cross-Encoders (ii) and Bi-Encoders \cite{humeau2019poly,thakur2020augmented,qu-etal-2021-rocketqa}. Cross-Encoders compute the full interaction between pairs of queries and documents and induce a joint representation for a query-document pair by means of cross-attention. Transformed representation of the query-document pair is then fed to a relevance classifier; the encoder and classifier parameters are updated jointly in an end-to-end fashion \cite{nogueira2019multi,macavaney2020teaching,Khattab2020colbert}. This paradigm is usually impractical for end-to-end ranking due to slow matching and retrieval. Recent work addresses this challenge by performing late interaction and by precomputing token-level representations \cite{Khattab2020colbert,gao-etal-2020-modularized}. Nonetheless, neural rankers are still predominantly used for re-ranking the top-ranked results returned by some base ranker. The alternative paradigm -- the so-called Bi-Encoders -- computes vector representations of documents and queries independently; it then relies on fast similarity computations in the vector space of precomputed query and document embeddings. All similarity-specialized multilingual encoders described in \S\ref{sec:sent-specialized-models} belong to this category of Bi-Encoders. \added[comment=In response to suggested related work; last citation is a book citation from Morgan Claypool]{Contrary to most NLP tasks, document-level ad-hoc IR deals with much longer text sequences. For instance, one notable approach computes document scores as an interpolation between a pre-ranking score and a weighed sum of scores of the top-k highest scoring sentences \cite{akkalyoncu-yilmaz-etal-2019-cross}. Our approach  scores local regions of documents independently (§\ref{sec:localrelmatching}); this is most similar to the BERT-MaxP model which encodes and scores individual passages of a document \cite{dai-sigir19}. For further discussion on long document matching we refer the reader to Chapter 3.3 of Lin et al.'s handbook \cite{lin2021pretrained}}. 

\added[comment=in response to reviewer 1 and the three paper references]{A related recent line of research targets \textit{cross-lingual transfer of (monolingual) rankers}, where such rankers are typically trained on English data and then applied in a monolingual non-English setting \cite{shi-etal-2020-cross,shi-etal-2021-cross,mrtydi}. This is different from our \textit{cross-lingual retrieval} evaluation setting where queries and documents are in different languages. A systematic comparative study focused on the suitability of the multilingual text encoders for \textit{diverse ad-hoc CLIR tasks} and language pairs is still lacking.} 

\vspace{1.8mm}
\noindent
\textbf{CLIR Evaluation and Application.} The cross-lingual ability of mBERT and XLM has been investigated by probing and analyzing their internals \cite{karthikeyan2019cross}, as well as in terms of downstream performance \cite{pires2019multilingual,wu2019beto}. In CLIR, these models as well as dedicated multilingual sentence encoders have been evaluated on tasks such as cross-lingual question-answer retrieval \cite{yang-etal-2020-multilingual-muse}, bitext mining \cite{ziemski2016united,ZWEIGENBAUM18.12-BUCC}, and semantic textual similarity (STS) \cite{hoogeveen2015,lei2016semi}. Yet, the models have been primarily evaluated on sentence-level retrieval, while classic ad-hoc (unsupervised) document-level CLIR has not been in focus. Further, previous work has not provided a large-scale comparative study across diverse language pairs and with different model variants, nor has tried to understand and analyze the differences between sentence-level and document-level tasks, or the impact of domain versus language transfer. In this work, we aim to fill these gaps. 







%% file: 3-models.tex
\section{Multilingual Text Encoders}
\label{s:mencoders}

We first provide an overview of all pretrained multilingual models in our evaluation. We discuss general-purpose multilingual text encoders (\S\ref{sec:mbert-xlm}), as well as specialized multilingual sentence encoders in \S\ref{sec:sent-specialized-models}. Finally, we describe the supervised rankers based on multilingual encoders (\S\ref{sec:l2r}). For completeness, we first briefly describe the baseline CLIR model based on CLWEs (\S\ref{sec:clwes}). 

\subsection{CLIR with (Static) Cross-lingual Word Embeddings}
\label{sec:clwes}

We assume a query $Q_{L_1}$ issued in a source language $L_1$, and a document collection of $N$ documents $D_{i, L_2}$, $i=1,\ldots,N$ in a target language $L_2$. Let $d=\{t_1,t_2,\dots,t_{|D|}\} \in D$ be a document with $|D|$ terms $t_i$. CLIR with static CLWEs represents queries and documents as vectors $\overrightarrow{Q},\overrightarrow{D}\in \mathbb{R}^d$ in a $d$-dimensional shared embedding space \cite{vulic2015sigir,litschko2019evaluating}. Each term is represented independently with a pre-computed static embedding vector $\overrightarrow{t_i} = emb\left(t_i\right)$. There exist a range of methods for inducing shared embedding spaces with different levels of supervision, such as parallel sentences, comparable documents, small bilingual dictionaries, or even methods without any supervision \cite{ruder2019survey}. Given the shared CLWE space, both query and document representations are obtained as aggregations of their term embeddings. We follow Litschko et al. \cite{litschko2019evaluating} and represent documents as the weighted sum of their terms' vectors, where each term's weight corresponds to its inverse document frequency (idf): \footnote{Only document term embeddings are idf-scaled.}

\begin{equation}
\label{eq:static_wemb_eq}
\overrightarrow{d} = \sum_{i = 1}^{N_d}{\mathit{idf}(t^d_i) \cdot \overrightarrow{t^d_i}}    
\end{equation} 

\noindent Documents are then ranked in decreasing order of the cosine similarity between their embeddings and the query embedding.

\subsection{Multilingual (Transformer-Based) Language Models: mBERT and XLM}
\label{sec:mbert-xlm}

\begin{figure*}
    \begin{center}
    \includegraphics[width=\linewidth]{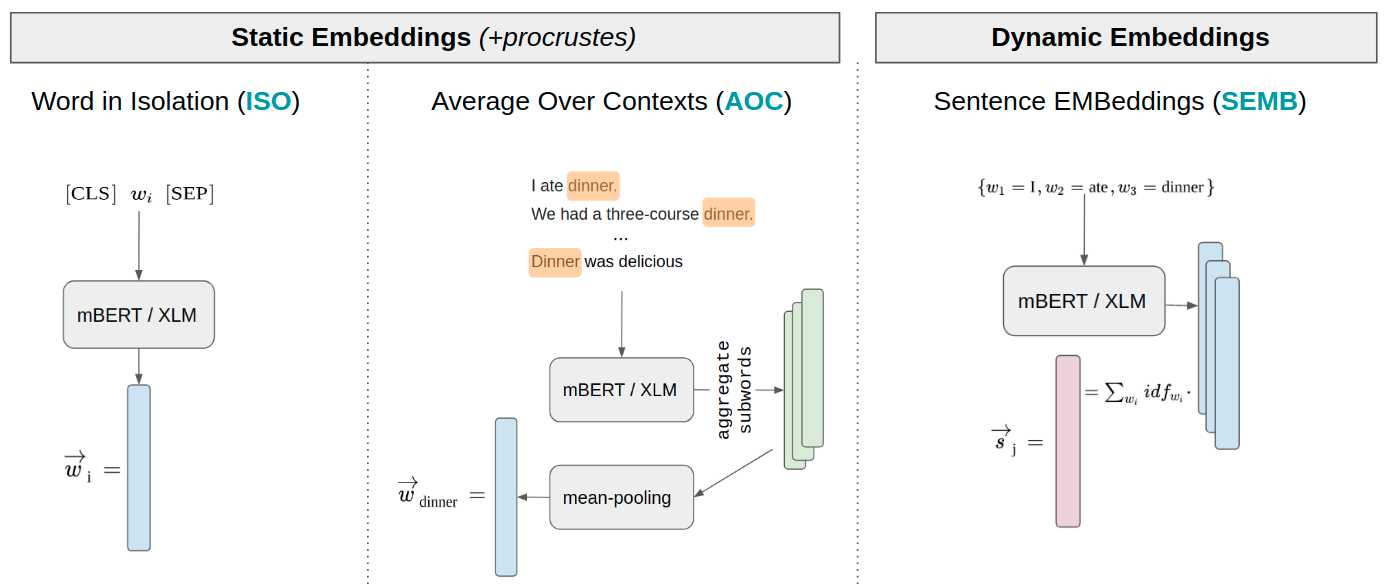}
    \caption{CLIR Models based on Multilingual Transformers. \textbf{Left:} Induce a static embedding space by encoding each vocabulary term in isolation; then refine the bilingual space for a specific language pair using the standard Procrustes projection. \textbf{Middle:} Aggregate different contextual representations of the same vocabulary term to induce static embedding space; then refine the bilingual space for a specific language pair using the standard Procrustes projection. \textbf{Right:} Direct encoding of a query-document pair with the multilingual encoder.}
    \label{fig:overview}
    \end{center}
    \vspace{-2mm}
\end{figure*}

Massively multilingual pretrained neural language models such as mBERT and XLM(-R) can be used as a dynamic embedding layer to produce contextualized word representations, since they share a common input space on the subword level (e.g. word-pieces, byte-pair-encodings) across all languages. Let us assume that a term (i.e., a word-level token) is tokenized into a sequence of $K$ subword tokens ($K\geq 1$; for simplicity, we assume that the subwords are word-pieces (\textit{wp})): $t_i=\big\{\textit{wp}_{i,k}\big\}^{K}_{k=1}$. The  multilingual encoder then produces contextualized subword embeddings for the term's $K$ constituent subwords $\overrightarrow{wp_{i,k}}$, $k=1,\ldots,K$, and we can aggregate these subword embeddings to obtain the representation of the term $t_i$: $\overrightarrow{t_i} = \psi\left(\{\overrightarrow{wp_{i,k}}\}^K_{k = 1}\right)$, where the function $\psi()$ is the aggregation function over the $K$ constituent subword embeddings. Once these term embeddings $\overrightarrow{t_i}$ are obtained, we follow the same CLIR setup as with CLWEs in \S\ref{sec:clwes}. We illustrate three different approaches for obtaining word and sentence representations from pretrained transformers in Figure~\ref{fig:overview} and describe them in more detail in what follows.
%
    
%

\vspace{1.8mm}

\vspace{1.4mm}
\noindent \textbf{Static Word Embeddings from Multilingual Transformers.} We first use multilingual transformers (mBERT and XLM) in two different ways to induce static word embedding spaces for all languages. In a simpler variant, we feed terms into the encoders \textit{in isolation} (\textbf{ISO}), that is, without providing any surrounding context for the terms. This effectively constructs a static word embedding table similar to what is done in \S\ref{sec:clwes}, and allows the CLIR model (\S\ref{sec:clwes}) to operate at a non-contextual word level. 
An empirical CLIR comparison between ISO and CLIR operating on traditionally induced CLWEs \cite{litschko2019evaluating} then effectively quantifies how well multilingual encoders (mBERT and XLM) capture word-level representations \cite{vulic2020probing}. 


In the second, more elaborate variant we do leverage the contexts in which the terms appear, constructing \textit{average-over-contexts} embeddings (\textbf{AOC}). For each term $t$ we collect a set of sentences $s_i \in \mathcal{S}_t$ in which the term $t$ occurs. We use the full set of Wikipedia sentences $\mathcal{S}$ to sample sets of contexts $\mathcal{S}_t$ for each vocabulary term $t$. For a given sentence $s_i$ let $j$ denote the position of $t$'s first occurrence. We then transform $s_i$ with mBERT or XLM as the encoder, $enc(s_i)$, and extract the contextualized embedding of $t$ via \textit{mean-pooling}, i.e., by averaging embeddings of its constituent subwords, $\psi\left(\{\overrightarrow{wp_{j,k}}\}^K_{k = 1}\right) = 1/K \cdot \sum_{k = 1}^{K}{\overrightarrow{wp_{j,k}}}$. Here, the function $\psi()$ is implemented as \textit{mean-pooling}, i.e., we obtain the contextualized representation of the term as the average of contextualized vectors of its constituent subwords. For each vocabulary term, we obtain $N_t = min(|\mathcal{S}_t|,\tau)$ contextualized vectors, with $|\mathcal{S}_t|$ as the number of Wikipedia sentences containing $t$ and $\tau$ as the maximal number of sentence samples for a term. The final static embedding of $t$ is then simply the average over the $N_t$ contextualized vectors. 

The obtained static AOC and ISO embeddings, despite being induced with multilingual encoders, however, did not appear to be lexically well-aligned across languages \cite{Liu:2019conll,cao2019multilingual}. We evaluated the static ISO and AOC embeddings induced for different languages with multilingual encoders (mBERT and XLM), on the bilingual lexicon induction (BLI) task \cite{glavas-etal-2019-properly}. We observed poor BLI performance, suggesting that further projection-based alignment of respective monolingual ISO and AOC spaces is warranted.    
To this end, we adopted the standard Procrustes method \cite{smith2017offline,Artetxe:2018acl} for learning an orthogonal linear projection from the embedding (sub)space of one language to the embedding space of the other language \cite{glavas-etal-2019-properly}. Let $D = \{(w^k_{L1}, w^k_{L2})\}^{K}_{k = 1}$ be the word translation dictionary between the two languages $L1$ and $L2$, containing $K$ word translation pairs. Let $\mathbf{X}_S = \{\mathbf{x}^k_{L1}\}^{K}_{k = 1}$ and $\mathbf{X}_T = \{\mathbf{x}^k_{L2}\}^{K}_{k = 1}$ be row-aligned matrices containing stacked embeddings of $\{w^k_{L1}\}^{K}_{k = 1}$ and $\{w^k_{L2}\}^{K}_{k = 1}$, respectively. We then obtain the projection matrix $\mathbf{W}$ by minimizing the Euclidean distance between the projection of $\mathbf{X}_S$ and the target matrix $\mathbf{X}_T$ \cite{mikolov2013exploiting}: $\mathbf{W} = \argmin_{\mathbf{W}}\lVert \mathbf{X}_{L1} \mathbf{W} - \mathbf{X}_{L2} \rVert_2$. If we constrain $\mathbf{W}$ to be orthogonal, the above optimization problem becomes the famous Procrustes problem, with the following closed-form solution \cite{schonemann1966generalized}:

\vspace{-0.5em}

\begin{align}
    \mathbf{W} &= \mathbf{UV}^\top, \, \text{with} \notag \\
    \mathbf{U\Sigma V}^\top &= \mathit{SVD}(\mathbf{X}_{T} {\mathbf{X}_{S}}^\top).
    \label{eq:proc}
\end{align}


\noindent In our experiments, for each language pair, we always project the AOC (ISO) embeddings of the query language to the AOC (ISO) embedding space of the document collection language, using the learned projection matrix $\mathbf{W}$.

\vspace{1.8mm}
\noindent \textbf{Direct Text Embedding with Multilingual Transformers}. 
%
%
%
%
In both AOC and ISO, we exploit the multilingual (contextual) encoders to obtain the static embeddings for word types (i.e., terms): we can then leverage these static word embeddings obtained from contextualized encoders in exactly the same ad-hoc CLIR setup (\S\ref{sec:clwes}) in which CLWEs had previously been evaluated \cite{litschko2019evaluating}. In an arguably more straightforward approach, we also use pretrained multilingual Transformers (i.e., mBERT or XLM) to directly semantically encode the whole input text \added{similar to encoding sentences into  \textit{Sentence EMBeddings}} (\textbf{SEMB}).  
%
To this end, we encode the input text by averaging the contextualized representations of all terms in the text (we again compute the weighted average, where the terms' IDF scores are used as weights, see \S\ref{sec:clwes}). For SEMB, we take the contextualized representation of each term $t_i$ to be the contextualized representation of its first subword token, i.e., $\overrightarrow{t_i} = \psi\left(\{\overrightarrow{wp_{i,k}}\}^K_{k = 1}\right) = \overrightarrow{wp_{i,1}}.$\footnote{In our preliminary experiments taking the vector of the first term's subword consistently outperformed averaging vectors of all its constituent subwords.} 



\subsection{Specialized Multilingual Sentence Encoders}
\label{sec:sent-specialized-models}
Off-the-shelf multilingual Transformers (mBERT and XLM) have been shown to yield sub-par performance in unsupervised text similarity tasks; therefore, in order to be successful in semantic text (sentences or paragraph) comparisons, they first need to be fine-tuned on text matching (typically sentence matching) datasets \cite{reimers2020making-distil,cao2019multilingual,Zhao:2020acl}. 
Such encoders \textit{specialized for semantic similarity} are supposed to encode sentence meaning more accurately, supporting tasks that require unsupervised (ad-hoc) semantic text matching. In contrast to off-the-shelf mBERT and XLM, which contextualize (sub)word representations, these models directly produce a semantic embedding of the input text. We provide a brief overview of the models included in our comparative evaluation.

\vspace{1.4mm}
\noindent
\textbf{Language Agnostic SEntence Representations (LASER)} \cite{artetxe2019LASER} adopts a standard seque\-nce-to-sequence architecture typical for neural machine translation (MT). It is trained on 223M parallel sentences covering 93 languages. The encoder is a multi-layered bidirectional LSTM and the decoder is a single-layer unidirectional LSTM. The 1024-dimensional sentence embedding is produced by max-pooling over the outputs of the encoder's last layer. The decoder then takes the sentence embedding as additional input at each decoding step. The decoder-to-encoder attention and language identifiers on the encoder side are deliberately omitted, so that all relevant information gets `crammed' into the fixed-sized sentence embedding produced by the encoder. In our experiments, we directly use the output of the encoder to represent both queries and documents.


\vspace{1.4mm}
\noindent
\textbf{Multilingual Universal Sentence Encoder (m-USE)}
is a general purpose sentence embedding model for transfer learning and semantic text retrieval tasks \cite{yang-etal-2020-multilingual-muse}. It relies on a standard dual-encoder neural framework \cite{chidambaram-etal-2019-learning-dual-encoder-framework,ijcai2019-746} with shared weights, trained in a multi-task setting with an additional translation bridging task. For more details, we refer the reader to the original work. There are two pretrained m-USE instances available -- we opt for the 3-layer Transformer encoder with average-pooling.  

%


\vspace{1.4mm}
\noindent
\textbf{Language-agnostic BERT Sentence Embeddings (LaBSE)} \cite{feng2020language-labse} is another neural dual-encoder framework, also trained with parallel data. Unlike LASER and m-USE, where the encoders are trained from scratch on parallel data, LaBSE starts its training from a pretrained mBERT instance (i.e., a 12-layer Transformer network pretrained on the concatenated corpora of 100+ languages). In addition to the multi-task training objective of m-USE, LaBSE additionally uses standard self-supervised objectives used in pretraining of mBERT and XLM: masked and translation language modeling (MLM and TLM, see \S\ref{sec:rw}). 
For further details, we refer the reader to the original work.

%
%

\vspace{1.4mm}
\noindent \textbf{DISTIL} \cite{reimers2020making-distil} is a teacher-student framework for injecting the knowledge obtained through specialization for semantic similarity from a specialized monolingual transformer (e.g., BERT) into a non-specialized multilingual transformer (e.g., mBERT). It first specializes for semantic similarity a monolingual (English) teacher encoder $M$ using the available semantic sentence-matching datasets for supervision. In the second, \textit{knowledge distillation} step a pretrained multilingual student encoder $\widehat{M}$ is trained to mimic the output of the teacher model. 
For a given batch of sentence-translation pairs $ \mathcal{B} = \{(s_j, t_j)\}$, the teacher-student distillation training minimizes the following loss: 

{
\footnotesize
\begin{equation*}
\mathcal{J}(\mathcal{B}) = \frac{1}{|\mathcal{B}|} \sum_{j\in \mathcal{B}} \left[\left(M(s_j)-\widehat{M}(s_j)\right)^2 + \left(M(s_j)-\widehat{M}(t_j)\right)^2\right].
\end{equation*}
}
\noindent The teacher model $M$ is Sentence-BERT \cite{reimers2019sentence}, BERT specialized for embedding sentence meaning on semantic text similarity \cite{cer-etal-2017-semeval} and natural language inference \cite{williams2018broad} datasets. The teacher network only encodes English sentences $s_i$. The student model $\widehat{M}$ is then trained to produce for both $s_j$ and $t_j$ the same representation that $M$ produces for $s_j$.  
We benchmark different DISTIL models in our CLIR experiments, with the student $\widehat{M}$ initialized with different multilingual transformers.

\subsection{Learning to (Re-)Rank with Multilingual Encoders}
\label{sec:l2r}

Finally, we consider another common setup, in which some relevance judgments (typically in English) are available and can be leveraged as supervision for fine-tuning multilingual encoders for ad-hoc retrieval. 
We consider two common scenarios: (1) an abundance of relevance annotations from other retrieval tasks and collections (but none for the target collection on which we want to perform ad-hoc retrieval) and (2) a small number of relevance judgments for the target collection. 
As an example of the former, we apply pointwise rankers pretrained on large-scale data (and based on multilingual encoders) in document-level CLIR on the CLEF benchmark. For the latter, we use a small number of CLEF relevance judgments to fine-tune, via contrastive metric-based learning, the representation space of the multilingual encoder. These two fine-tuning approaches are described in what follows.         


\vspace{1.4mm}
\noindent
\textbf{Pointwise Ranking with Multilingual Transformers.} A common learning-to-rank (L2R) approach with pretrained neural text encoders is the pointwise classification of query-document pairs \cite{nogueira2019multi,macavaney2020teaching}. In this so-called Cross-Encoder approach, the input to the pretrained encoder is a query-document concatenation. More specifically, let query $q$ consist of the query (subword) tokens $t^q_1,\dots t^q_n$ and document $d$ consist of the document (subword) tokens $t^d_1,\dots t^d_m$. The input to the pretrained encoder is then \texttt{[CLS] $t^q_1,\dots t^q_n$ [SEP] $t^d_1,\dots t^d_m$ [SEP]}, with \texttt{[CLS]} and \texttt{[SEP]} being the special sequence start and segment separation tokens of the corresponding pretrained encoder, e.g., BERT \cite{devlin-etal-2019-bert}. When needed, the documents are truncated in order to meet the maximum input length constraint of the respective pretrained transformer.  
This setup -- i.e., concatenation of two texts -- is common for various sentence-pair classification tasks in natural language processing (e.g., natural language inference or semantic text similarity). The encoded representation of the sequence start token (\texttt{[CLS]}), taken from the last layer of the Transformer-based encoder is then fed into a feed-forward classifier with a single hidden layer, which outputs the probability of the document being relevant for the query. The parameters of the feed-forward classifier are (fine-)tuned together with the encoder's parameters in an end-to-end fashion, by means of minimizing the standard cross-entropy loss. The positive training instances are simply the available relevance judgments (i.e., queries paired with documents indicated as relevant); the non-trivial negative instances are commonly created by pairing queries with irrelevant documents that are ranked highly by some baseline ranker (e.g., BM25) \cite{nogueira2019multi}.  
%

Pointwise neural rankers have been shown both ineffective (many false positive) and inefficient (at inference, one has to feed the query paired with each document through the classifier) when used to rank the entire document collection from scratch. In contrast, they have been very successful in re-ranking the top of the ranking produced by some baseline ranker, such as BM25. In CLIR, however, due to the very limited lexical overlap between languages, one cannot use base rankers based on lexical overlap such as BM25 or the vector space model (VSM).
In our re-ranking experiments (see \S\ref{sec:reranking}) we thus employ our unsupervised CLIR rankers based on multilingual encoders from \S\ref{sec:sent-specialized-models} as base rankers. 

\vspace{1.4mm}
\noindent
\textbf{Contrastive Metric-Based Learning.} The above pointwise approach which \textit{cross-encodes} each query-document pair (by concatenating the query with each document and passing them jointly to the encoder) is computationally heavy. Therefore, as mentioned before, it is primarily used for re-ranking. Further, it introduces additional trainable parameters of the classifier: their reliable estimation requires a large amount of training instances. In contrast, in most ad-hoc retrieval setups, one at best has a handful of relevance judgments for the test collection of interest. An alternative approach in such low-supervision settings is to use the few available relevance judgments to reshape the representation space of the (multilingual) text encoder, without training a dedicated relevance classifier (i.e., no additional trainable parameters).        
%
In this so called Bi-Encoder paradigm, the objective is to bring representations of queries, produced independently by the pretrained encoder, closer to the representations of their relevant documents (produced again independently by the same encoder) than to the representations of irrelevant documents. The objectives of contrastive metric-based learning push the instances that stand in a particular relation (e.g., query and \textit{relevant} document) closer together according to a predefined similarity or distance metric (e.g., cosine similarity) than corresponding pairs that do not stand in the relation of interest (e.g., the same query and some irrelevant document). It is precisely the approach used for obtaining multilingual encoders specialized for sentence similarity tasks covered in \S\ref{sec:sent-specialized-models} \cite{reimers2019sentence,feng2020language-labse,yang-etal-2020-multilingual-muse}. 

We propose to use contrastive metric-based learning to fine-tune the representation space for the concrete ad-hoc retrieval task, using a limited amount of relevance judgments available for the target collection. To this end, we employ a popular contrastive learning objective referred to as Multiple Negative Ranking Loss (MNRL) \cite{thakur2020augmented}. Given a query vector $q_i$, a relevant document $d_i^+$ and a set of in-batch negatives $\{d^-_{i,j}\}^m_{j=1}$ we fine-tune the parameters of a pretrained multilingual encoder by minimizing MNRL, given as:

{
\footnotesize
\begin{equation*}
\mathcal{L}\left(q_i, d^+_i,\{d^-_{i,j}\}^m_{j=1}\right) = - \log \frac{e^{\lambda \cdot \text{sim}(q_i,d^+_i)}}{e^{\lambda \cdot \text{sim}(q_i,d^+_i)} + \sum_{j=1}^m e^{\lambda \cdot \text{sim}(q_i,d^-_{i,j})}} \end{equation*}
}%
\noindent Each document, the relevant $d_j^+$ and each of the irrelevant $d^-_{i,j}$, receives a score that reflects their similarity to the query $q_i$: for this, we rely on cosine similarity, i.e. $\text{sim}(q_i,d_j) = \text{cos}(q_i,d_j)$. Document scores, scaled with a temperature factor $\lambda$, are then converted into a probability distribution with a softmax function. The loss is then, intuitively, the negative log likelihood of the relevant document $d_j^+$. In \S\ref{sec:finetuningdistilmbert}, we fine-tune in this manner the best-performing multilingual encoder (see \S\ref{sec:base_results_docs}).


%% file: 4-unsup.tex
\section{Unsupervised CLIR}
\label{sec:experintalsetup}

We first present the experiments demonstrating the suitability of pretrained multilingual models as text encoders for ad-hoc unsupervised CLIR (i.e., we evaluate models described in \S\ref{sec:mbert-xlm} and \S\ref{sec:sent-specialized-models}). 

\subsection{Experimental Setup}
\label{sec:exp_setup_unsup}

\textbf{Evaluation Data.} We follow the experimental setup of Litschko et al.\ \cite{litschko2019evaluating}, and compare the models from \S\ref{s:mencoders} on language pairs comprising five languages: English (EN), German (DE), Italian (IT), Finnish (FI) and Russian (RU). For document-level retrieval we run experiments for the following nine language pairs: EN-\{FI, DE, IT, RU\}, DE-\{FI, IT, RU\}, FI-\{IT, RU\}. We use the 2003 portion of the CLEF benchmark \cite{braschler2003clef},\footnote{\url{http://catalog.elra.info/en-us/repository/browse/ELRA-E0008/}} with 60 queries per language pair. 
\added[comment=Table 1 shows for Europarl 100k docs for each target language - actual numbers are slightly below because empty lines are filtered - should I include them here?]{For sentence-level retrieval}, also following \cite{litschko2019evaluating}, for each language pair we sample from Europarl \cite{Koehn:2005} 1K source language sentences as queries and 100K target language sentences as the ``document collection''. \added[comment=Feel free to improve table format]{We refer the reader to Table~\ref{tbl:stats} for } \added[comment=Table 6 has also doc freq statis - is this one here redundant?]{summary} \added[comment=EN-/RU-rows not included in Europarl table because RU is not in Europarl and we don't test on English]{statistics}.\footnote{Russian is not included in Europarl and we therefore exclude it from sentence-level experiments. Further, since some multilingual encoders have not seen Finnish data in pretraining, we additionally report the results over a subset of language pairs that do not involve Finnish.}

\setlength{\tabcolsep}{10.2pt}
\begin{table}[t]
\centering
\caption{\added{Basic statistics of CLEF 2003 and Europarl test collections: number of documents (\#doc); average number of tokens produced by the XLM/mBERT tokenizer (\#xlm, \#mbert); average number of relevant documents per query (\#rel).}}
\vspace{1mm}
{\footnotesize
{
\begin{tabularx}{\linewidth}{c cccc cccc}
\toprule
& \multicolumn{4}{c}{CLEF 2003} & \multicolumn{4}{c}{Europarl} \\ 
\cmidrule(lr){2-5} \cmidrule(lr){6-9} 
Lang. & \#doc & \#rel & \#mbert & \#xlm & \#doc & \#rel & \#mbert & \#xlm \\ \cmidrule(lr){1-9}
EN & 169k & 18.6 & 700.4 & 746.6 & - & - & - & - \\
DE & 295k & 32.6 & 490.9 & 518.7 & 100k & 1 & 35.6 & 38.3 \\
IT & 158k & 15.9 & 482.8 & 491.8 & 100k & 1 & 41.4 & 38.2 \\
FI & 55k & 10.7 & 648.7 & 623.7 & 100k & 1 & 37.6 & 38.1 \\
RU & 17k & 5.4 & 557.8 & 536.3 & - & - & - & - \\ 
\bottomrule
\end{tabularx} 
}}
\label{tbl:stats}
\end{table}

\vspace{1.8mm}
\noindent
\textbf{Baseline Models.} In order to establish whether multilingual encoders outperform CLWEs in a fair comparison, we compare their performance against the strongest CLWE-based CLIR model from the recent comparative study \cite{litschko2019evaluating}, dubbed Proc-B. Proc-B induces a bilingual CLWE space from pretrained monolingual \textsc{fastText} embeddings\footnote{\url{https://fasttext.cc/docs/en/pretrained-vectors.html}} using the linear projection computed as the solution of the Procrustes problem given the dictionary of word-translation pairs. Compared to simple Procrustes mapping, Proc-B iteratively (1) augments the word translation dictionary by finding mutual nearest neighbours and (2) induces a new projection matrix using the augmented dictionary. The final bilingual CLWE space is then plugged into the CLIR model from \S\ref{sec:clwes}.   

Our document-level retrieval SEMB models do not get to see the whole document but only the first $128$ word-piece tokens. For a more direct comparison, we therefore additionally evaluate the Proc-B baseline (Proc-B\textsubscript{LEN}) which is exposed to exactly the same amount of document text as the multilingual XLM encoder (i.e., the leading document text corresponding to first $128$ word-piece tokens) 
Finally, we compare CLIR models based on multilingual Transformers to  a baseline relying on machine translation baseline (MT-IR). In MT-IR, 1) we translate the query to the document language using Google Translate and then 2) perform monolingual retrieval using a standard Query Likelihood Model \cite{ponte1998language} with Dirichlet smoothing \cite{zhai2004study}. 


\vspace{1.8mm}
\noindent
\textbf{Model Details.} For all multilingual encoders we experiment with different input sequence lengths: $64$, $128$, $256$ subword tokens. 
For AOC we collect (at most) $\tau=60$ contexts for each vocabulary term: for a term not present at all in Wikipedia, we fall back to the ISO embedding of that term. We also investigate the impact of $\tau$ in \S\ref{sec:discussion}. In all cases (SEMB, ISO, AOC), we surround the input with the special sequence start and end tokens of respective pretrained models: $[CLS]$ and $[SEP]$ for BERT-based models and $\langle s\rangle$ and $\langle/s \rangle$ for XLM-based models. For vanilla multilingual encoders (mBERT and XLM) and all three variants (SEMB, ISO, AOC), we independently evaluate representations from different Transformer layers (cf. \S\ref{sec:discussion}). For comparability, for ISO and AOC -- methods that effectively induce static word embeddings using multilingual contextual encoders -- we opt for exactly the same term vocabularies used by the Proc-B baseline, namely the top 100K most frequent terms from respective monolingual fastText vocabularies. 
%
We additionally experiment with three different instances of the DISTIL model: (i) $\text{DISTIL}_{\text{XLM-R}}$ initializes the student model with the pretrained XLM-R transformer \cite{conneau2019unsupervised}; $\text{DISTIL}_{\text{USE}}$ instantiates the student as the pretrained m-USE instance \cite{yang-etal-2020-multilingual-muse}; whereas $\text{DISTIL}_{\text{DistilmBERT}}$ distils the knowledge from the Sentence-BERT teacher into a multilingual version of DistilBERT \cite{sanh2019distilbert}, a 6-layer transformer pre-distilled from mBERT.\footnote{Working with mBERT directly instead of its distilled version led to similar scores, while increasing running times.} For SEMB models we scale embeddings of special tokens (sequence start and end tokens, e.g., \texttt{[CLS]} and \texttt{[SEP]} for mBERT) with the mean IDF value of input terms. 


\subsection{Document-Level CLIR Results}
\label{sec:base_results_docs}

\begin{table*}[t!]
\centering
\caption{Document-level CLIR results (Mean Average Precision, MAP). \textbf{Bold}: best model for each language-pair. *: difference in performance w.r.t. Proc-B significant at $p = 0.05$, computed via paired two-tailed t-test with Bonferroni correction.}
\vspace{1mm}
\def\arraystretch{0.95}
{\scriptsize
{
\begin{tabularx}{\linewidth}{l X X X X X X X X X X X} 
\toprule
 & EN-FI & EN-IT & EN-RU & EN-DE & DE-FI & DE-IT & DE-RU & FI-IT & FI-RU & AVG & w/o FI \\ \midrule
 \textit{Baselines} \\\midrule
MT-IR & .276 & \textbf{.428} & .383 & \textbf{.263} & \textbf{.332} & \textbf{.431} & .238 & \textbf{.406} & .261 & \textbf{.335} & \textbf{.349} \\
Proc-B & .258 & .265 & .166 & .288 & .294 & .230 & .155 & .151 & .136 & .216 & .227 \\
$\text{Proc-B}_{\text{LEN}}$ & .165 & .232 & .176 & .194 & .207 & .186 & .192 & .126 & .154 & .181 & .196 \\ \midrule
\multicolumn{12}{l}{\textit{Models based on multilingual Transformers}} \\ \midrule
$\text{SEMB}_{\text{XLM}}$ & .199* & .187* & .183 & .126* & .156* & .166* & .228 & .186* & .139 & .174 & .178 \\
$\text{SEMB}_{\text{mBERT}}$ & .145* & .146* & .167 & .107* & .151* & .116* & .149* & .117 & .128* & .136 & .137 \\ \cdashline{1-12}[.4pt/1pt]
$\text{AOC}_{\text{XLM}}$  & .168 & .261 & .208 & .206* & .183 & .190 & .162 & .123 & .099 & .178 & .206 \\ 
$\text{AOC}_{\text{mBERT}}$ & .172* & .209* & .167 & .193* & .131* & .143* & .143 & .104 & .132 & .155 & .171 \\ \cdashline{1-12}[.4pt/1pt]
$\text{ISO}_{\text{XLM}}$  & .058* & .159* & .050* & .096* & .026* & .077* & .035* & .050* & .055* & .067 & .083 \\ 
$\text{ISO}_{\text{mBERT}}$  & .075* & .209 & .096* & .157* & .061* & .107* & .025* & .051* & .014* & .088 & .119 \\ \midrule
\multicolumn{9}{l}{\textit{Similarity-specialized sentence encoders (with parallel data supervision)}} \\ \midrule 
$\text{DISTIL}_{\text{FILTER}}$ & .291 & .261 & .278 & .255 & .272 & .217 & .237 & .221 & .270 & .256 & .250 \\ \cdashline{1-12}[.4pt/1pt]
$\text{DISTIL}_{\text{XLM-R}}$  & .216 & .190* & .179 & .114* & .237 & .181 & .173 & .166 & .138 & .177 & .167 \\
$\text{DISTIL}_{\text{USE}}$ & .141* & .346* & .182 & .258 & .139* & .324* & .179 & .104 & .111 & .198 & .258 \\
$\text{DISTIL}_{\text{DistilmBERT}}$ & \textbf{.294} & .290* & \textbf{.313} & .247* & .300 & .267* & \textbf{.284} & .221* & \textbf{.302}* & .280 & .280 \\ \cdashline{1-12}[.4pt/1pt]
LaBSE & .180* & .175* & .128 & .059* & .178* & .160* & .113* & .126 & .149 & .141 & .127 \\
LASER & .142 & .134* & .076 & .046* & .163* & .140* & .065* & .144 & .107 & .113 & .094 \\
m-USE & .109* & .328* & .214 & .230* & .107* & .294* & .204 & .073 & .090 & .183 & .254 \\
\bottomrule
\end{tabularx}
}}
\label{tab:clefselfsup}
\vspace{-1.5mm}
\end{table*}

We show the performance (MAP) of multilingual encoders on document-level CLIR tasks in Table~\ref{tab:clefselfsup}.
The first main finding is that none of the self-supervised models (mBERT and XLM in ISO, AOC, and SEMB variants) outperforms the CLWE baseline Proc-B. However, the full Proc-B baseline has, unlike mBERT and XLM variants, been exposed to the full content of the documents. A fairer comparison, against Proc-B\textsubscript{LEN}, which has also been exposed only to the first $128$ tokens, reveals that SEMB and AOC variants come reasonably close, albeit still do not outperform Proc-B\textsubscript{LEN}. This suggests that the document-level retrieval could benefit from encoders able to encode longer portions of text, e.g., \cite{beltagy2020longformer,zaheer2020big}. For document-level CLIR, however, these models would first have to be ported to multilingual setups. Scaling embeddings by their \textit{idf} (Proc-B) effectively filters out high-frequent terms such as stopwords. We therefore experiment with explicit a priori stopword filtering in $\text{DISTIL}_\text{DistilmBERT}$, dubbed $\text{DISTIL}_\text{FILTER}$. Results show that performance deteriorates which indicates that stopwords provide important contextualization information. While SEMB and AOC variants exhibit similar performance, ISO variants perform much worse. The direct comparison between ISO and AOC demonstrates the importance of contextual information and seemingly limited usability of off-the-shelf multilingual encoders as word encoders, if no context is available, and if they are not further specialized to encode word-level information \cite{liu2021fast}. 


Similarity-specialized multilingual encoders, which rely on pretraining with parallel data, yield mixed results. Three models, $\text{DISTIL}_\text{DistilmBERT}$, $\text{DISTIL}_\text{USE}$ and m-USE, generally outperform the Proc-B baseline.\footnote{As expected, m-USE and $\text{DISTIL}_\text{USE}$ perform poorly on language pairs involving Finnish, as they have not been trained on any Finnish data.} 
LASER is the only encoder trained on parallel data that does not beat the Proc-B baseline. We believe this is because (a) LASER's recurrent encoder provides text embeddings of lower quality than Transfor\-mer-based encoders of m-USE and DISTIL variants and (b) it has not been subjected to any self-supervised pretraining like DISTIL models. Even the best-performing CLIR model based on a multilingual encoder ($\text{DISTIL}_\text{DistilmBERT}$) overall falls behind the MT-based baseline (MT-IR). However, it is very important to note that the performance of MT-IR critically depends on the quality of MT for the concrete language pair: for language pairs with weaker MT (e.g., FI-RU, EN-FI, FI-RU, DE-RU), $\text{DISTIL}_\text{DistilmBERT}$ can substantially outperform MT-IR (e.g., 9 MAP points for FI-RU and DE-RU). In contrast, the gap in favor of MT-IR is, as expected, largest for the pairs of large typologically similar languages, for which also the most reliable MT systems exist: EN-IT, EN-DE. In other words, the feasibility and robustness of a strong MT-IR CLIR model seems to diminish with more distant language pairs and lower-resource languages. 

The variation in results with similarity-specialized sentence encoders indicates that: (a) despite their seemingly similar high-level architectures typically based on dual-encoder networks \cite{cer2018universal}, it is important to carefully choose a sentence encoder in document-level retrieval, and (b) there is an inherent mismatch between the granularity of information encoded by the current state-of-the-art text representation models and the document-level CLIR task.



\subsection{Sentence-Level Cross-Lingual Retrieval}
\label{sec:sentence-level-clir}

\begin{table*}[!t]
\centering
\caption{Sentence-level CLIR results (MAP). \textbf{Bold}: best model for each language-pair. *: difference in performance with respect to Proc-B, significant at $p = 0.05$, computed via paired two-tailed t-test with Bonferroni correction.}
\vspace{1mm}
\def\arraystretch{0.93}
{\scriptsize
{
\begin{tabularx}{\linewidth}{l X X X X X X X X}
\toprule
& EN-FI & EN-IT & EN-DE & DE-FI & DE-IT & FI-IT & AVG & w/o FI \\\midrule
\textit{Baselines} \\\midrule
MT-IR & .659 & .803 & .725 & .541 & .694 & .698 & .687 & .740 \\
Proc-B & .143 & .523 & .415 & .162 & .342 & .137 & .287 & .427 \\\midrule
\multicolumn{9}{l}{\textit{Models based on multilingual Transformers}}\\ \midrule
$\text{SEMB}_{\text{XLM}}$ & .309* & .677* & .465 & .391* & .495* & .346* & .447 & .545\\
$\text{SEMB}_{\text{mBERT}}$ & .199* & .570 & .355 & .231* & .481* & .353* & .365 & .469 \\ \cdashline{1-9}[.4pt/1pt]
$\text{AOC}_{\text{XLM}}$  & .099 & .527 & .274* & .102* & .282 & .070* & .226 & .361 \\
$\text{AOC}_{\text{mBERT}}$ & .095* & .433* & .274* & .088* & .230* & .059* & .197 & .312 \\ \cdashline{1-9}[.4pt/1pt]
$\text{ISO}_{\text{XLM}}$  & .016* & .178* & .053* & .006* & .017* & .002* & .045 & .082 \\ 
$\text{ISO}_{\text{mBERT}}$  & .010* & .141* & .087* & .005* & .017* & .000* & .043 & .082 \\ \midrule
\multicolumn{9}{l}{\textit\textit{Similarity-specialized sentence encoders (with parallel data supervision)}} \\ \midrule
$\text{DISTIL}_{\text{XLM-R}}$ & .935* & .944* & .943* & .911* & .919* & .914* & .928 & .935 \\
$\text{DISTIL}_{\text{USE}}$ & .084* & .960* & .952* & .137 & .920* & .072* & .521 & .944 \\
$\text{DISTIL}_{\text{DistilmBERT}}$ & .847* & .901* & .901* & .811* & .842* & .793* & .849 & .882 \\ \cdashline{1-9}[.4pt/1pt]
LaBSE & .971* & .972* & .964* & .948* & .954* & .951* & .960 & .963 \\
LASER\qquad\qquad & \textbf{.974*} & \textbf{.976*} & \textbf{.969*} & \textbf{.967*} & \textbf{.965*} & \textbf{.961*} & \textbf{.969} & \textbf{.970} \\
m-USE & .079* & .951* & .929* & .086* & .886* & .039* & .495 & .922 \\ \bottomrule
\end{tabularx}
}}
\label{tab:europarl-selfsupervised}
\vspace{-1.5mm}
\end{table*}

We show the sentence-level CLIR performance in Table~\ref{tab:europarl-selfsupervised}. 
Unlike in the document-level CLIR task, self-supervised SEMB variants here manage to outperform Proc-B. The better relative SEMB performance than in document-level retrieval is somewhat expected: sentences are much shorter than documents (i.e., typically shorter than the maximal sequence length of $128$ word pieces). All purely self-supervised mBERT and XLM variants, however, perform worse than the translation-based baseline. 


Multilingual sentence encoders specialized with parallel data excel in sentence-level CLIR, all of them substantially outperforming the competitive MT-IR baseline. This however, does not come as much of a surprise, since these models (a) have been trained using parallel data (i.e., sentence translations), and (b) have been optimized exactly on the sentence similarity task. In other words, in the context of the cross-lingual sentence-level task, these models are effectively supervised models.  
The effect of supervision is most strongly pronounced for LASER, which was, being also trained on parallel data from Europarl, effectively subjected to in-domain training. We note that at the same time LASER was the weakest model from this group on average in the document-level CLIR task. 

The fact that similarity-specialized multilingual encoders perform much better in sentence-level than in document-level CLIR suggests viability of a different approach to document-level retrieval: instead of obtaining a single encoding for the document, one may (independently) encode its sentences (or larger windows of content) and (independently) measure their semantic correspondence to the query. We investigate this \textit{localized relevance matching} approach to document-level CLIR with similarity-specialized multilingual encoders in the next section (\S\ref{sec:localrelmatching}).             


\subsection{Localized Relevance Matching}
\label{sec:localrelmatching}

Contrary to most NLP tasks, in ad-hoc document retrieval we face the challenge of semantically representing long documents. According to \cite{robertson1994some}, documents can be viewed either as a concatenation of topically heterogeneous short sub-documents (\textit{``Scope Hypothesis''}) or as a more verbose version of a short document on the same topic (\textit{``Verbosity Hypothesis''}). Under both hypotheses, the source of relevance of the document for the query is localized, i.e., there should exist (at least one) segment (relatively short w.r.t. the length of the whole document) that is the source of relevance of the document for the query. 
Furthermore, a query may represent an information need on a specific aspect of a topic that is simply not discussed at the beginning, but rather somewhere later in the document: the maximum input sequence length imposed by neural text encoders directly limits the retrieval effectiveness in such cases. Even if we assume that we can encode the complete document with our multilingual encoders, these document representations would likely become semantically less precise (i.e., fuzzier) as they would aggregate contextualized representations of many more tokens; in \S\ref{sec:discussion} we validate this empirically and show that simply increasing the maximum sequence length of multilingual encoders does not improve their retrieval performance.  

Recent work proposed pretraining procedures for encoding long documents \cite{zaheer2020big,dai2019transformer-xl,beltagy2020longformer}. These models have been pretrained only for English. Pretraining their multilingual counterparts, however, would require extremely large and massively multilingual corpora and computational resources of the scale that we do not have at our disposal. 
%
%
In the following, we instead experiment with two resource-lean alternatives: we represent documents either as (1) sets of overlapping text \textit{segments} obtained by running a sliding window over the document or (2) a collection of document \textit{sentences}, which we then encode independently similar to \cite{akkalyoncu-yilmaz-etal-2019-cross}. For a single document, we now need to store multiple semantic representations (i.e., embeddings), one for each text segment or sentence. While these approaches clearly increase the index size as well as the retrieval latency (as the query representation needs to be compared against embeddings of all document segments or sentences), sufficiently fast ad-hoc retrieval for most use cases can still be achieved with highly efficient approximate search libraries such as FAISS \cite{Johnson-2017faiss}. Representing documents as multiple segments or sentences allows for fine-grained local matching against the query: a setting in which sentence-specialized multilingual encoders are supposed to excel, see Table \ref{tab:europarl-selfsupervised}.  

\begin{table*}[t!]
\centering
\caption{Document-level CLIR results for \textit{localized relevance matching} against document \textit{segments} (overlapping 128-token segments). Document relevance is the average of relevance scores of $k$ highest-scoring segments. Results (for 9 language pairs from CLEF) shown for the Proc-B baseline and all similarity-specialized encoders. $\Delta$ AVG denotes relative performance increases/decreases w.r.t. the respective base performances from Table \ref{tab:clefselfsup}.}
\vspace{1mm}
\def\arraystretch{0.95}
{
{\fontsize{6.5pt}{6.5pt}\selectfont
\begin{tabularx}{\linewidth}{l X X X X X X X X X X X r} 
\toprule
& k & EN-FI & EN-IT & EN-RU & EN-DE & DE-FI & DE-IT & DE-RU & FI-IT & FI-RU & AVG & $\Delta$ AVG\\
\midrule
\multirow{4}{*}{Proc-B} & 1 & .242 & .253 & .182 & .286 & .280 & .217 & .158 & .147 & .166 & .215 & $-$0.86  \\
& 2 & .241 & .244 & .153 & .287 & .282 & .207 & .116 & .147 & .115 & .199 & $-$2.40  \\
& 3 & .234 & .235 & .150 & .277 & .269 & .194 & .113 & .153 & .109 & .193 & $-$3.04 \\
& 4 & .228 & .217 & .135 & .255 & .276 & .171 & .105 & .167 & .098 & .184 & $-$3.95 \\ \cdashline{1-13}[.4pt/1pt]
\multirow{4}{*}{$\text{DISTIL}_{\text{DmBERT}}$} & 1 & .330 & .327 & .248 & .365 & .324 & .293 & \textbf{.244} & .268 & .236 & .293 & +1.32 \\
& 2 & \textbf{.349} & .315 & \textbf{.269} & .382 & \textbf{.347} & .287 & .216 & \textbf{.272} & \textbf{.226} & \textbf{.296} & +1.61 \\
& 3 & .323 & .291 & .261 & .353 & .335 & .268 & .226 & .248 & .208 & .279 & $-$0.03 \\
& 4 & .299 & .263 & .207 & .330 & .316 & .236 & .189 & .217 & .181 & .249 & $-$3.10 \\ \cdashline{1-13}[.4pt/1pt]
\multirow{4}{*}{$\text{DISTIL}_{\text{XLM-R}}$} & 1 & .284 & .218 & .160 & .233 & .267 & .195 & .162 & .181 & .156 & .206 & +2.92 \\
& 2 & .279 & .208 & .164 & .253 & .264 & .194 & .179 & .187 & .157 & .209 & +3.25 \\
& 3 & .264 & .191 & .141 & .228 & .253 & .188 & .145 & .171 & .157 & .193 & +1.60 \\
& 4 & .236 & .169 & .105 & .203 & .237 & .167 & .114 & .153 & .113 & .166 & $-$1.07 \\ \cdashline{1-13}[.4pt/1pt]
\multirow{4}{*}{$\text{DISTIL}_{\text{USE}}$} & 1 & .149 & .355 & .202 & .363 & .138 & .332 & .199 & .074 & .118 & .214 & +1.64 \\
& 2 & .162 & \textbf{.377} & .192 & .416 & .136 & \textbf{.344} & .197 & .081 & .095 & .222 & +2.42 \\
& 3 & .150 & .344 & .180 & .391 & .137 & .319 & .181 & .079 & .091 & .208 & +1.00 \\
& 4 & .135 & .313 & .163 & .364 & .128 & .280 & .158 & .064 & .086 & .188  & $-$1.03 \\ \cdashline{1-13}[.4pt/1pt]
\multirow{4}{*}{LaBSE} & 2 & .212 & .118 & .102 & .189 & .199 & .103 & .060 & .085 & .083 & .128 & $-$1.13 \\
& 1 & .221 & .108 & .124 & .141 & .198 & .093 & .077 & .063 & .143 & .130 & $-$1.32 \\
& 3 & .198 & .104 & .080 & .153 & .190 & .089 & .052 & .076 & .066 & .112 & $-$2.90 \\
& 4 & .186 & .088 & .065 & .128 & .176 & .075 & .036 & .069 & .049 & .097 & $-$4.42 \\ \cdashline{1-13}[.4pt/1pt]
\multirow{4}{*}{mUSE} & 1 & .073 & .345 & .215 & .361 & .082 & .331 & .210 & .053 & .084 & .195 & +1.19 \\
& 2 & .102 & .370 & .213 & \textbf{.404} & .085 & \textbf{.344} & .209 & .056 & .085 & .208 & +2.46 \\
& 3 & .083 & .333 & .198 & .376 & .074 & .296 & .186 & .053 & .082 & .187 & +0.38 \\
& 4 & .075 & .291 & .178 & .348 & .067 & .257 & .178 & .047 & .077 & .169 & $-$1.43 \\ \cdashline{1-13}[.4pt/1pt]
\multirow{4}{*}{LASER} & 1 & .135 & .058 & .049 & .075 & .155 & .054 & .070 & .082 & .061 & .082 & +1.40 \\
& 2 & .150 & .069 & .071 & .099 & .161 & .055 & .060 & .088 & .062 & .091 & +2.26 \\
& 3 & .136 & .054 & .053 & .074 & .142 & .044 & .052 & .072 & .049 & .075 & +0.71 \\
& 4 & .113 & .037 & .038 & .057 & .118 & .032 & .045 & .052 & .038 & .059 & $-$0.91 \\ 
\bottomrule
\end{tabularx}
}}
\label{tab:chunking}
\vspace{-1.5mm}
\end{table*}

\vspace{1.8mm}
\noindent\textbf{Localized Relevance Matching: Segments.} In this approach, we slide a window of size 128 word tokens over the document with a stride of 42 tokens, creating multiple overlapping 128-word segments from the input document. Each segment is then encoded separately, leveraging the encoders from \S\ref{s:mencoders}. We then score for relevance each segment by comparing its respective embedding with the query embedding. We then compute the final relevance score by averaging the relevance scores of the top-$k$ highest-scoring segments. 

Table \ref{tab:chunking} displays the results of all multilingual encoders in our comparison, for $k \in \{1, 2, 3, 4\}$.\footnote{For $k = 1$, the relevance of the document is exactly the score of the highest scoring segment.} For most encoders (with the exception of LaBSE and the Proc-B baseline) we observe gains from segment-based localized relevance matching: we observe the largest average gain of 3.25 MAP points for $\text{DISTIL}_{\text{XLM-R}}$ (from 0.177 for document encoding to 0.209 for segment-based localized relevance matching). Most importantly, we observe gains for our best-performing multilingual encoder $\text{DISTIL}_{\text{DmBERT}}$: localized relevance matching (for $k=2$) pushes its performance by 1.6 MAP points (the base performance of 0.28 is shown in Table \ref{tab:clefselfsup}). We suspect that applying IDF-Sum in Proc-B (see \S\ref{sec:clwes}) has a similar (albeit query-independent) soft filtering effect to localized relevance matching and that this is why localized relevance matching does not yield any gains for this competitive baseline. 

\begin{figure}
    \centering
    \includegraphics[scale=0.33]{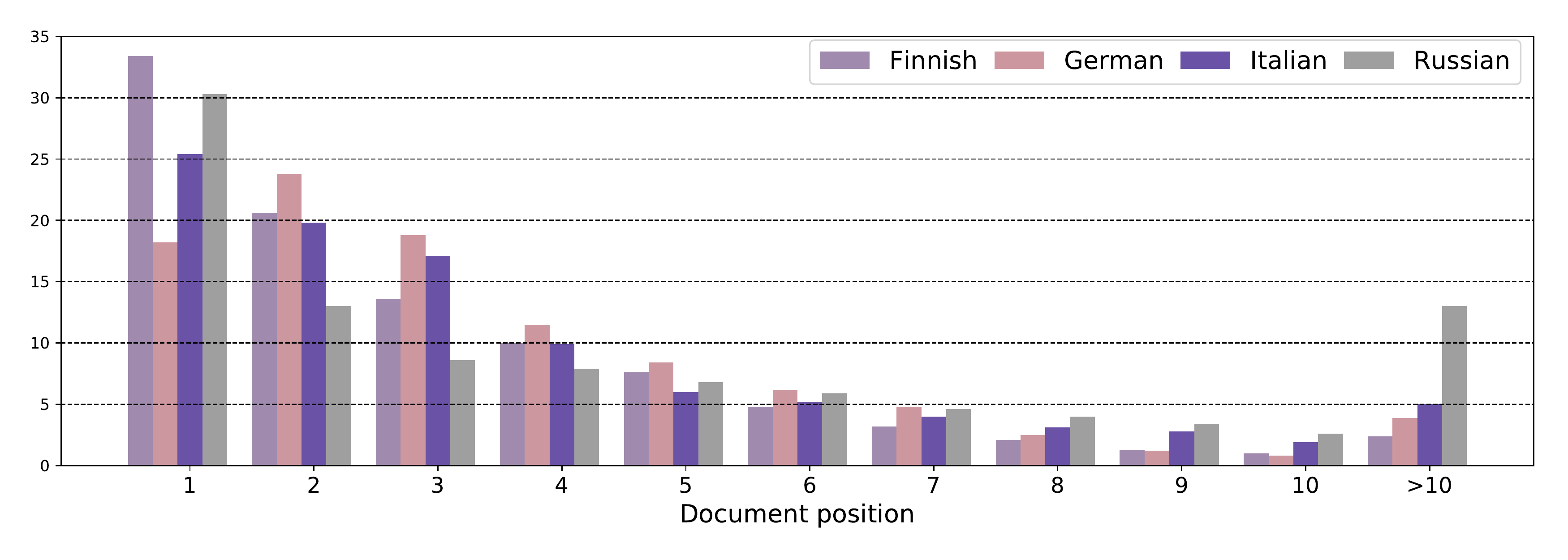}
    \caption{Comparison of within-document positions of top-ranked segments in segment-based localized relevance matching for different \textit{collection languages}. Proportions aggregated across all multilingual CLIR models from Table \ref{tab:chunking}.}
    \label{fig:segment_positions_langs}
\end{figure}

For all five multilingual encoders for which we observe gains from localized relevance matching, these gains are the largest for $k = 2$, i.e., when we average the relevance scores of the two highest-scoring segments. In 63.7\% of the cases, the two highest-scoring segments are mutually consecutive, overlapping segments: we speculate that in those cases it is the span of text in which they overlap that contains the signal that makes the document relevant for the query. \added[comment=in response to reviewer request; do I need to say \text{inter alia} in the citation? I'm sure there's more work with the same insight]{These findings are in line with similar observations from previous work \cite{akkalyoncu-yilmaz-etal-2019-cross,dai-sigir19}: aggregating local relevance signals yields strong retrieval results.}
Matching queries with the most similar segment embedding effectively filters out the rest of the document. Our results suggest that improvements are mostly consistent across language pairs: we only fail to observe gains when Russian is the language of the target document collection. Localized relevance matching can in principle decrease the performance if segmentation produces (many) false positives (i.e., irrelevant segments with high semantic similarity with the query). We suspect this to more often be the case for Russian than for the other languages. We further investigate this by comparing positions of high-scoring segments across document collection languages. 
We look at the distributions of document positions among the top-ranked 100 segments (gathered from all collection documents): the distributions of top-ranked segment results per positions in respective documents (i.e., $1$ indicates the first segment of the document, $2$ the second, etc.) are shown for each of the four collection languages (aggregated across all multilingual encoders from Table~\ref{tab:chunking}) in Figure~\ref{fig:segment_positions_langs}. 
%
%
The distributions of positions of high-scoring segments confirms our suspicion that something is different for Russian compared to other languages: we observe a much larger presence of high-scoring segments that appear later in the documents, i.e., at positions larger than 10 ($>$10): while there is between 2\% and 5\% of such ``late'' high-scoring segments in Italian, German, and Finnish collections, in the Russian collection there is 13\% of such segments. Our manual inspection confirmed that these late segments are indeed most often false positives (i.e., irrelevant for the query, yet with representations highly similar to those of the queries): this presumably causes the lower performance on *-RU benchmarks.  

\begin{figure}
    \centering
    \includegraphics[scale=0.33]{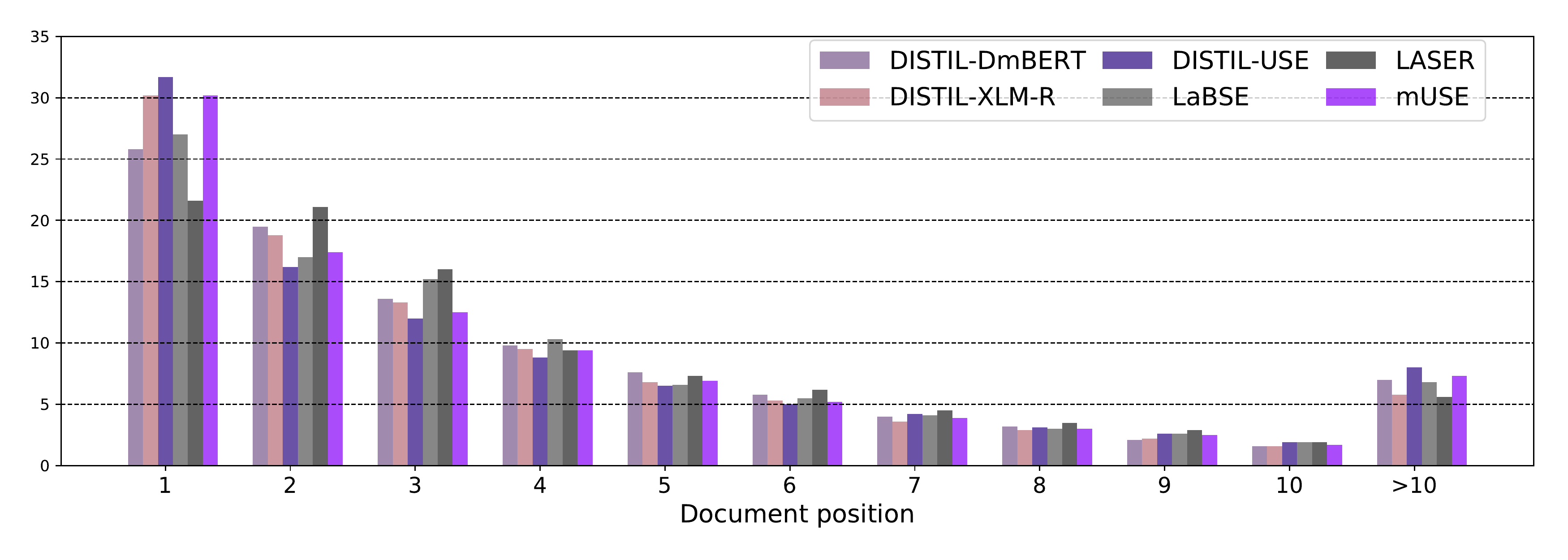}
    \caption{Comparison of within-document positions of top-ranked segments in segment-based localized relevance matching for different \textit{multilingual text encoders}. Proportions aggregated across all multilingual CLIR models from Table \ref{tab:chunking}.}
    \label{fig:segment_positions_encoders}
\end{figure}

Figure~\ref{fig:segment_positions_encoders} compares the individual multilingual encoders along the same dimension: document positions of the segments they rank the highest. Unlike for collection languages, we do not observe major differences across multilingual encoders -- for all of them, the top-ranked segments seem to have similar within-document position distributions, with  ``early'' segments (positions $1$ and $2$) having the highest relative participation at the top of the ranking. In general, the analysis of positions of high-scoring segments empirically validates the intuition that the most relevant content is often localized at the beginning of the target documents within the newswire CLEF corpora, which in turn reflects the writing style of the news domain. 

\vspace{1.8mm}
\noindent\textbf{Localized Relevance Matching: Sentences.} The selection of the segmentation strategy can have a profound effect on the effectiveness of localized relevance matching. Instead of (overlapping 128-token) segments, one could, for example, measure the relevance of each document sentence for the query and (max-)pool the sentence relevance scores. Sentence-level segmentation and relevance pooling is particularly interesting when considering multilingual encoders that have been specialized precisely for sentence-level semantics (i.e., produce accurate sentence-level representations; see \S\ref{sec:sent-specialized-models}). In Table \ref{tab:splitting} we show the results of sentence-level localized relevance matching for all multilingual encoders. Unlike with segment-based localized relevance matching (see Table \ref{tab:chunking}), here we see improvements for all multilingual encoders: what is more important, improvements over the baseline performance of the same encoders (see Table \ref{tab:clefselfsup}) are substantially larger than for segment-based localized relevance matching (e.g., 10 and 3.8 MAP-point improvements from sentence matching for LASER and LaBSE, respectively, compared to 2-point improvement for LASER and an 1-point MAP drop for LaBSE from segment matching). Sentence-level matching with the best-performing base multilingual encoder $\text{DISTIL}_{\text{DmBERT}}$ and pooling over two highest-ranking sentences (i.e., $k = 2$) yields the best unsupervised CLIR score that we observed overall (31.4 MAP points). For all encoders, averaging the scores of $k = 2$ or $k = 3$ highest-scoring sentences gives better results than considering only the single best sentence (i.e., $k = 1$) -- this would indicate that the query-relevant content is still not overly localized within documents (i.e., not confined to a single sentence).        

%

\begin{table*}[t!]
\centering
\caption{Document-level CLIR results for \textit{localized relevance matching} against document \textit{sentences}. Document relevance is the average of relevance scores of $k$ highest-scoring sentences. Results (for 9 language pairs from CLEF) shown for the Proc-B baseline and all multilingual encoders specialized for encoding sentence-level semantics. $\Delta$ AVG denotes relative performance increases/decreases w.r.t. the respective base performances from Table \ref{tab:clefselfsup}.}
\vspace{1mm}
\def\arraystretch{0.95}
{
{\fontsize{6.5pt}{6.5pt}\selectfont
\begin{tabularx}{\linewidth}{l X X X X X X X X X X X r} 
\toprule
& k & EN-FI & EN-IT & EN-RU & EN-DE & DE-FI & DE-IT & DE-RU & FI-IT & FI-RU & AVG & $\Delta$ AVG \\ \midrule
\multirow{4}{*}{Proc-B} & 1 & .219 & .207 & .136 & .191 & .235 & .203 & .138 & .089 & .126 & .171 & $-$5.16 \\
& 2 & .216 & .273 & .158 & .238 & .267 & .247 & .176 & .142 & .122 & .204 & $-$1.90 \\
& 3 & .229 & .267 & .165 & .245 & .284 & .231 & .168 & .153 & .120 & .207 & $-$1.61 \\
& 4 & .231 & .247 & .173 & .235 & .286 & .215 & .166 & .150 & .120 & .202 & $-$2.07 \\ \cdashline{1-13}[.4pt/1pt]
\multirow{4}{*}{$\text{DISTIL}_{\text{DmBERT}}$} & 1 & \textbf{.381} & .288 & .249 & .332 & .338 & .248 & .234 & .234 & .234 & .282 & $+$0.24 \\
& 2 & .371 & .313 & \textbf{.303} & .399 & .343 & .285 & .286 & .246 & \textbf{.280} & \textbf{.314} & $+$3.44 \\
& 3 & .360 & .308 & .288 & .407 & \textbf{.359} & .274 & \textbf{.288} & .247 & .279 & .312 & $+$3.26 \\
& 4 & .345 & .298 & .264 & .382 & .352 & .262 & .263 & \textbf{.248} & .271 & .298 & $+$1.87 \\ \cdashline{1-13}[.4pt/1pt]
\multirow{4}{*}{$\text{DISTIL}_{\text{XLM-R}}$} & 1 & .323 & .220 & .144 & .239 & .316 & .215 & .148 & .200 & .149 & .217 & $+$4.00 \\
& 2 & .339 & .250 & .199 & .306 & .305 & .246 & .200 & .229 & .196 & .252 & $+$7.51 \\
& 3 & .328 & .260 & .205 & .311 & .318 & .237 & .209 & .222 & .208 & .255 & $+$7.81 \\
& 4 & .311 & .263 & .188 & .298 & .319 & .225 & .178 & .220 & .179 & .242 & $+$6.52 \\ \cdashline{1-13}[.4pt/1pt]
\multirow{4}{*}{$\text{DISTIL}_{\text{USE}}$} & 1 & .131 & .270 & .181 & .332 & .121 & .244 & .200 & .070 & .054 & .178 & $-$2.01 \\
& 2 & .139 & .331 & .226 & .408 & .134 & .321 & .240 & .076 & .132 & .223 & $+$2.50 \\
& 3 & .131 & .329 & .220 & \textbf{.433} & .129 & .334 & .235 & .074 & .129 & .224 & $+$2.56 \\
& 4 & .134 & .340 & .212 & .428 & .122 & .329 & .225 & .068 & .124 & .220 & $+$2.21 \\ \cdashline{1-13}[.4pt/1pt]
\multirow{4}{*}{LaBSE} & 1 & .188 & .182 & .126 & .167 & .185 & .147 & .101 & .112 & .112 & .147 & $+$0.57 \\
& 2 & .225 & .197 & .182 & .213 & .227 & .180 & .108 & .138 & .139 & .179 & $+$3.77 \\
& 3 & .245 & .186 & .157 & .234 & .255 & .163 & .089 & .136 & .110 & .175 & $+$3.39 \\
& 4 & .249 & .192 & .117 & .235 & .248 & .139 & .077 & .145 & .106 & .167 & $+$2.65 \\ \cdashline{1-13}[.4pt/1pt]
\multirow{4}{*}{mUSE} & 1 & .123 & .270 & .147 & .317 & .112 & .256 & .124 & .070 & .034 & .161 & $-$2.17 \\
& 2 & .139 & .368 & .212 & .395 & .127 & .334 & .187 & .079 & .069 & .212 & $+$2.92 \\
& 3 & .142 & \textbf{.369} & .230 & .428 & .122 & \textbf{.341} & .189 & .083 & .077 & .220 & $+$3.72 \\
& 4 & .138 & .357 & .220 & .429 & .116 & .331 & .172 & .081 & .086 & .214 & $+$3.13 \\ \cdashline{1-13}[.4pt/1pt]
\multirow{4}{*}{LASER} & 1 & .207 & .130 & .096 & .147 & .206 & .123 & .107 & .141 & .112 & .141 & $+$7.30 \\
& 2 & .175 & .172 & .127 & .184 & .206 & .138 & .133 & .165 & .129 & .159 & $+$9.07 \\
& 3 & .191 & .177 & .153 & .185 & .197 & .141 & .154 & .172 & .136 & .167 & $+$9.94 \\
& 4 & .175 & .172 & .133 & .179 & .184 & .131 & .125 & .166 & .123 & .154 & $+$8.60 \\
\bottomrule
\end{tabularx}
}}
\label{tab:splitting}
\vspace{-1.5mm}
\end{table*}

Finally, it is important to note that the gains in retrieval effectiveness (i.e., MAP gains) obtained with localized relevance matching (segment-level and sentence-level) come at the expense of reduced retrieval efficiency (i.e., increased retrieval time): the query representation now needs to be compared with each of the segment or sentence representations, instead of with only one aggregate representation for the whole document. The slowdown factor is proportional to the average number of segments/sentences per document in the document collection. Table~\ref{tab:numberofdocuments} summarizes the approximate slowdown factors (i.e., average numbers of segments and sentences) for CLEF document collections in different languages.         

\setlength{\tabcolsep}{14.5pt}
\begin{table*}[t!]
\centering
\caption{Increase in computational complexity (i.e., decrease in retrieval efficiency) due to localized relevance matching via segments and sentences.}
\vspace{1mm}
{\footnotesize
{
\begin{tabularx}{\linewidth}{l r r r r r r} 
\toprule
& & \multicolumn{2}{c}{Segmentation} & \multicolumn{2}{c}{Sentence Splitting} \\
& \#Documents & \#Segments & Factor & \#Sentences & Factor \\ \midrule
DE & 294,809 & 1,281,993 & 4.35 & 5,385,103 & 18.27 \\
IT & 157,558 & 749,855 & 4.76 &  2,225,069 & 14.12 \\
FI & 55,344 & 224,390 & 4.05 & 1,286,702 & 23.25 \\
RU & 16,715 & 72,102 & 4.31 & 289,740 & 17.33 \\
\bottomrule
\end{tabularx}
}}
\label{tab:numberofdocuments}
\end{table*}

\subsection{Further Analysis}
\label{sec:discussion}
We now further investigate three aspects that may impact CLIR performance of multilingual encoders: (1) layer(s) from which we take vector representations, (2) number of contexts used in AOC variants, and (3) sequence length in document-level CLIR.

\begin{figure}[!t]
\centering\includegraphics[width=\textwidth]{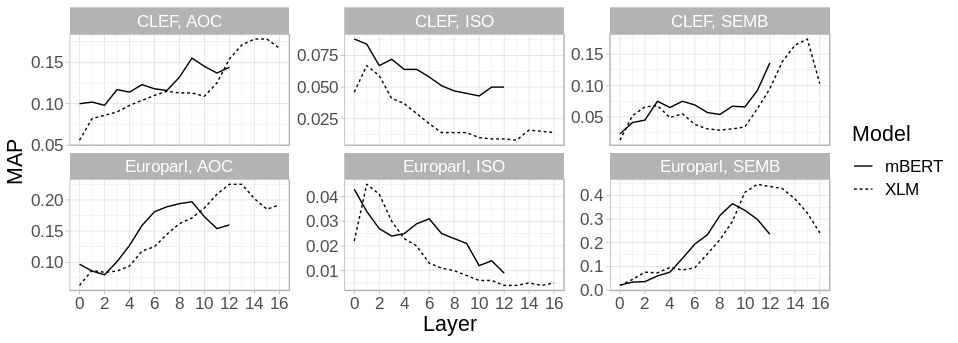}
\caption{CLIR performance of mBERT and XLM as a function of the Transformer layer from which we obtain the representations. Results (averaged over all language pairs) shown for all three encoding strategies (SEMB, AOC, ISO).}
\label{fig:layerplots}
\end{figure}

\vspace{1.8mm}
\noindent \textbf{Layer Selection.} All multilingual encoders have multiple layers and one may in principle choose to take (sub)word representations for CLIR at the output of any of them.    
Figure~\ref{fig:layerplots} shows the impact of taking subword representations after each layer for self-supervised mBERT and XLM variants. We find that the optimal layer differs across the encoding strategies (AOC, ISO, and SEMB; cf.~\S\ref{sec:mbert-xlm}) and tasks (document-level vs. sentence-level CLIR). ISO, where we feed the terms into encoders without any context, seems to do best if we take the representations from lowest layers. This makes intuitive sense, as the parameters of higher Transformer layers encode compositional rather than lexical semantics \cite{Ethayarajh:2019emnlp,rogers2020primer}. For AOC and SEMB, where both models obtain representations by contextualizing (sub)words in a sentence, we get the best performance for higher layers -- the optimal layers for document-level retrieval (L9/L12 for mBERT, and L15 for XLM) seem to be higher than for sentence-level retrieval (L9 for mBERT and L11/L12 for XLM). 

\begin{figure}[!t]
    \centering
    \includegraphics[width=\textwidth]{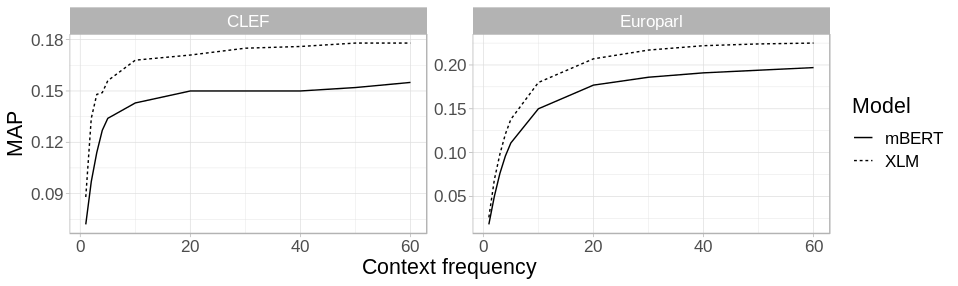}
    \caption{CLIR performance of AOC variants (mBERT and XLM) w.r.t. the number of contexts used to obtain the term embeddings.}
    \label{fig:contextplots}
    \vspace{-2mm}
\end{figure}

\vspace{1.8mm}
\noindent\textbf{Number of Contexts in AOC.} We construct AOC term embeddings by averaging contextualized representations of the same term obtained from different Wikipedia contexts. This raises an obvious question of a sufficient number of contexts needed for a reliable (static) term embedding. Figure~\ref{fig:contextplots} shows the AOC results depending on the number of contexts used to induce the term vectors (cf. $\tau$ in \S\ref{s:mencoders}). The AOC performance seems to plateau rather early -- at around 30 and 40 contexts for mBERT and XLM, respectively. Encoding more than 60 contexts (as we do in our main experiments) would therefore bring only negligible performance gains.  

\vspace{1.8mm}
\noindent\textbf{Input Sequence Length.} Multilingual encoders have a limited input length and they, unlike CLIR models operating on static embeddings (Proc-B, as well as our AOC and ISO variants), effectively truncate long documents. This limitation was, in part, also the motivation for localized relevance matching approaches from the previous section.
In our main experiments we truncated the documents to first $128$ word pieces. Now we quantify (Table~\ref{tbl:seqlen}) if and to which extent this has a detrimental effect on document-level CLIR performance.  
Somewhat counterintuitively, encoding a longer chunk of documents ($256$ word pieces) yields a minor performance deterioration (compared to the length of $128$) for \textit{all} multilingual encoders. We suspect that this is a combination of two effects: (1) it is more difficult to semantically accurately encode a longer portion of text, which leads to semantically less precise embeddings of $256$-token sequences; and (2) for documents in which the query-relevant content is not within the first $128$ tokens, that content might often also appear beyond the first $256$ tokens, rendering the increase in input length inconsequential to the recognition of such documents as relevant. These results, combined with gains obtained from localized relevance matching in the previous section render localized matching (i.e., document relevance pooled from segment- or sentence-level relevance scores) as a more promising strategy for retrieving long documents than attempts to increase the input length of multilingual transformers. Our findings from localized relevance matching seem to indicate that the relevance signal is highly localized: in such a setting, aggregating representations of very many tokens (i.e., across the whole document), e.g., with long-input transformers \cite{beltagy2020longformer,zaheer2020big}, is poised to produce semantically fuzzier (i.e., less precise) representations, from which it is harder to judge the document relevance for the query.     


\setlength{\tabcolsep}{2.5pt}
\begin{table*}[!t]
\centering
\caption{Document-level unsupervised CLIR results w.r.t. the input text length. Scores averaged over all language pairs not involving Finnish.}
\vspace{1mm}
\def\arraystretch{0.99}
{\scriptsize
{
\begin{tabularx}{\linewidth}{l X X X X X X X X}
\toprule
Length & $\text{SEMB}_\text{mBERT}$ & $\text{SEMB}_\text{XLM}$ & $\text{DIST}_\text{use}$ & $\text{DIST}_\text{XLM-R}$ & $\text{DIST}_\text{DmBERT}$ & mUSE & LaBSE & LASER \\ \midrule
64 & .104 & .128 & .235 & .167 & .237 & .254 & .127 & .089 \\
128 & .137 & .178 & .258 & .162 & .280 & .247 & .125 & .068 \\
256 & .117 & .158 & .230 & .146 & .250 & .197 & .096 & .027 \\ \bottomrule
\end{tabularx}
}}
\label{tbl:seqlen}
\end{table*}

\noindent 



%% file: 5-l2r.tex
\section{Supervised (Re-)Ranking}

We next evaluate, on the same document-level collection from CLEF, the CLIR effectiveness of the multilingual encoders that have been exposed to some amount of supervision, i.e., fine-tuned using certain amount of relevance judgments, described in \S\ref{sec:l2r}. We first discuss in \S\ref{sec:reranking} the performance of pointwise (re-)rankers based on mBERT trained on large-scale out-of-domain collections; we then analyse (\S\ref{sec:finetuningdistilmbert}) how contrastive in-domain fine-tuning affects CLIR performance. In both cases, we exploit annotated English data for model fine-tuning: the transfer to other languages is directly enabled by the multilingual nature of the encoders.       

\subsection{Re-ranking with Pointwise Rankers}
\label{sec:reranking}

\begin{table*}[t!]
\centering
\caption{Document-level CLIR results on the CLEF collection obtained by language and domain transfer of supervised re-ranking models. For each query, we re-rank the top 100 results produced by the base multilingual ranker with two mBERT-based L2R models trained on English data: MS MARCO \cite{msmarcodataset} (middle part of the table) and TREC ROBUST (bottom third of the table) \cite{trecrobust04,macavaney2020teaching}. \textbf{Bold}: the best performance in each column (for each language pairs and the average).}
\vspace{1mm}
\def\arraystretch{0.95}
{\scriptsize
{
\begin{tabularx}{\linewidth}{l X X X  X  X  X  X  X  X  X  X  X X} 
\toprule
 & EN-FI & EN-IT & EN-RU & EN-DE & DE-FI & DE-IT & DE-RU & FI-IT & FI-RU & AVG & $\Delta$ \\ \midrule
\textit{No re-ranking (reference)} \\
\midrule
Proc-B & .258 & .265 & .166 & .288 & .294 & .230 & .155 & .136 & .216 & .223 & -- \\ \cdashline{1-12}[.4pt/1pt]
$\text{DISTIL}_{\text{DmBERT}}$ & \textbf{.294} & .290 & \textbf{.313} & .247 & \textbf{.300} & .267 & \textbf{.284} & \textbf{.221} & \textbf{.302} & \textbf{.280} & --  \\
$\text{DISTIL}_{\text{XLM-R}}$ & .219 & .191 & .149 & .148 & .215 & .179 & .142 & .167 & .125 & .170 & --  \\
$\text{DISTIL}_{\text{USE}}$ & .141 & .346 & .182 & .258 & .139 & .324 & .179 & .104 & .111 & .198 & --  \\ \cdashline{1-12}[.4pt/1pt]
mUSE & .077 & .313 & .186 & .262 & .077 & .293 & .183 & .053 & .092 & .171 & --  \\
LaBSE & .191 & .163 & .136 & .087 & .172 & .136 & .103 & .117 & .140 & .138 & --  \\
LASER & .146 & .092 & .060 & .039 & .153 & .089 & .062 & .117 & .076 & .093 & --  \\ \midrule
\textit{Re-ranker trained on MS MARCO} \\\midrule
Proc-B & .269 & .283 & .250 & .297 & .248 & .214 & .220 & .157 & .151 & .232 & $+$0.90 \\ \cdashline{1-12}[.4pt/1pt]
$\text{DISTIL}_{\text{DmBERT}}$ & .282 & .319 & .252 & .279 & .257 & .263 & .261 & .210 & .202 & .258 & $-$2.13 \\
$\text{DISTIL}_{\text{XLM-R}}$ & .255 & .259 & .234 & .240 & .256 & .188 & .208 & .167 & .196 & .223 & $+$5.20 \\
$\text{DISTIL}_{\text{USE}}$ & .194 & .350 & .287 & .309 & .169 & .292 & .256 & .150 & .112 & .235 & $+$3.71 \\\cdashline{1-12}[.4pt/1pt]
mUSE & .135 & \textbf{.355} & .298 & .301 & .145 & \textbf{.295} & .260 & .113 & .101 & .223 & $+$5.19 \\
LaBSE & .252 & .274 & .215 & .160 & .248 & .209 & .169 & .152 & .170 & .205 & $+$6.71 \\
LASER & .195 & .185 & .128 & .100 & .194 & .167 & .160 & .128 & .135 & .155 & $+$6.21 \\ \midrule
\textit{Re-ranker trained on TREC ROBUST} \\ \midrule 
Proc-B & .290 & .292 & .141 & .310 & .278 & .214 & .148 & .108 & .103 & .209 & $-$1.38 \\ \cdashline{1-12}[.4pt/1pt]
$\text{DISTIL}_{\text{DmBERT}}$ & .284 & .283 & .153 & .274 & .252 & .246 & .130 & .147 & .119 & .210 & $-$6.98 \\
$\text{DISTIL}_{\text{XLM-R}}$ & .270 & .227 & .093 & .242 & .226 & .200 & .079 & .129 & .069 & .170 & $+$0.00 \\
$\text{DISTIL}_{\text{USE}}$ & .195 & .321 & .119 & .309 & .194 & .287 & .113 & .113 & .117 & .196 & $-$0.19 \\ \cdashline{1-12}[.4pt/1pt]
mUSE & .143 & .330 & .129 & \textbf{.313} & .139 & .261 & .131 & .086 & .079 & .179 & $+$0.82 \\
LaBSE & .275 & .234 & .086 & .158 & .245 & .180 & .076 & .115 & .077 & .161 & $+$2.25 \\
LASER & .201 & .164 & .121 & .095 & .171 & .137 & .118 & .111 & .093 & .135 & $+$4.19 \\
\bottomrule
\end{tabularx}
}}
\label{tab:clefrerank}
\vspace{-1.5mm}
\end{table*}

Transferring (re-)rankers across domains and/or languages is a promising method when in-language and in-domain fine-tuning data is scarce \cite{macavaney2019cedr}. We experimented with two pointwise rankers, both based on mBERT, pretrained on English relevance data. The first model\footnote{\url{https://huggingface.co/amberoad/bert-multilingual-passage-reranking-msmarco}} was trained on the large-scale MS MARCO passage retrieval dataset \cite{msmarcodataset}, consisting of approx. 400M tuples, each consisting of a query, a relevant passage and a non-relevant passage. Transferring rankers trained on MS MAR\-CO to various ad-hoc IR settings (i.e., domains) has been shown successful \cite{li2020parade,macavaney-etal-2020-sledge,craswell2021ms}. Here, we investigate the performance of this supervised ranker trained on MS MARCO in simultaneous domain and language transfer.  
The second multilingual pointwise ranker \cite{macavaney2020teaching} is trained on TREC 2004 Robust dataset \cite{trecrobust04}. Although TREC 2004 Robust is substantially smaller than MS MARCO (528K documents and 311K relevance judgments), by covering newswire documents it is domain-wise closer to our target CLEF test collection.   
As discussed in \S\ref{sec:l2r}, pointwise neural rankers are typically used to re-rank the top of the ranking produced by some base ranker, rather than to rank the whole collection from scratch. Accordingly, we use the two above-described mBERT-based pointwise re-rankers to re-rank the top $100$ documents from the initial rankings produced by each of the similarity-specialized multilingual encoders from \S\ref{sec:sent-specialized-models}).\footnote{We also experimented with re-ranking top $1,000$ documents, but the results were slightly worse for all base multilingual encoders than when re-ranking only the top 100 results.} 

Table \ref{tab:clefrerank} summarizes the results of our domain and language transfer experiments with the two pointwise mBERT-based re-rankers. For clarity, at the top of the table, we repeat the reference unsupervised CLIR performance of the similarity-specialized multilingual encoders (i.e., without any re-ranking) from Table \ref{tab:clefselfsup}.  
Intuitively, re-ranking -- both with the MS MARCO-trained model and TREC-trained model -- brings the largest gains for the weakest unsupervised rankers: mUSE, LaBSE, and LASER. The gains are somewhat larger when transferring the model trained on MS MARCO. However, re-ranking the results of the best-performing unsupervised ranker -- $\text{DISTIL}_{\text{DmBERT}}$ -- brings no performance gains; in fact, re-ranking with the TREC-trained model reduces the quality of the base ranking by 7 MAP points.            %
%
%
The transfer performance of the better-performing MS MARCO re-ranker in our CLIR benchmarks from CLEF depends on (1) the performance of the base ranker and (2) the target language pair. MS MARCO re-ranker improves the performance of our best-performing initial ranker, $\text{DISTIL}_{\text{DmBERT}}$, only for EN-DE and EN-IT, two language pairs in our evaluation for which the query language (EN) and collection language (DE, IT) are the closest to the source language of MS MARCO (EN) on which the re-ranker was trained; conversely, the MS MARCO re-ranking yields the largest performance drop for FI-RU, i.e., the pair of languages in our evaluation that are typologically most distant from EN. These results suggest that, assuming a strong multilingual encoder as the base ranker, supervised re-ranking does not transfer well to distant language pairs.  
Overall, our results are in line with the most recent findings from \cite{craswell2021ms}, which also suggests that a ranker trained only on the large dataset like MS MARCO (i.e., without any fine-tuning on the target collection) yields mixed ad-hoc retrieval results.

\subsection{Contrastive In-Domain Fine-Tuning}
\label{sec:finetuningdistilmbert}

\begin{figure}
    \centering
    \includegraphics[scale = 0.33]{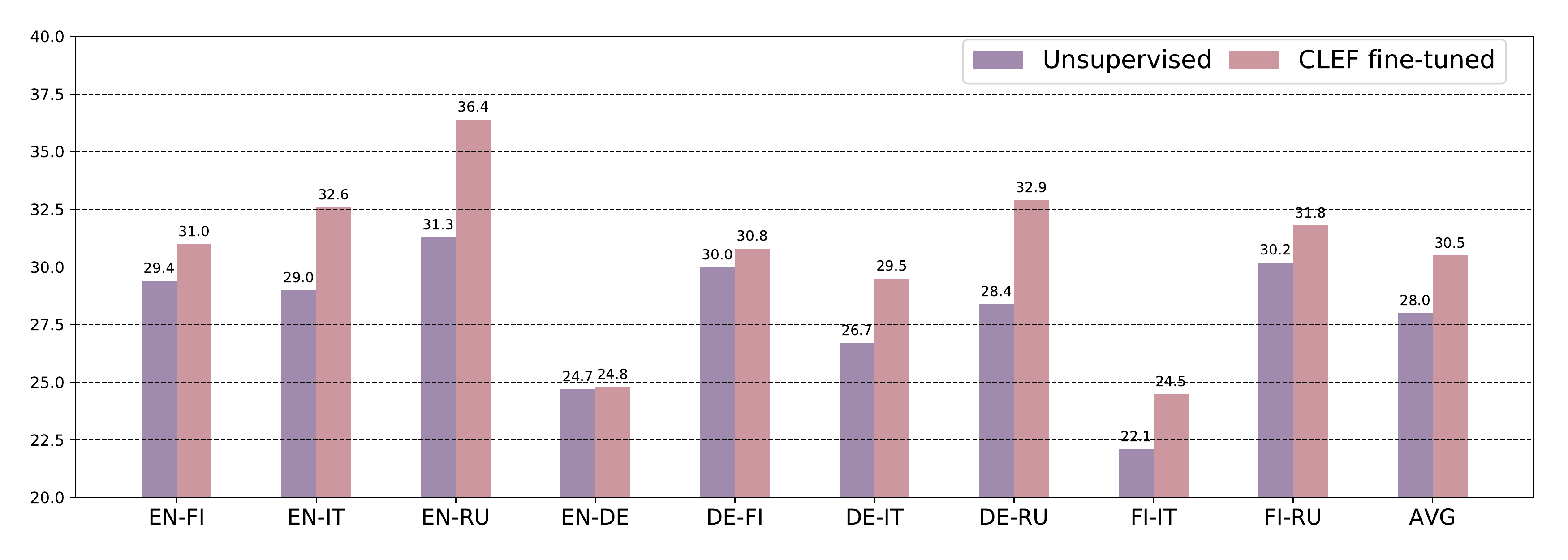}
    \caption{The effects of ``in-domain'' fine-tuning: comparison of CLIR performance with $\text{DISTIL}_{\text{DmBERT}}$ on the CLEF CLIR collections: (a) without any fine-tuning (i.e., an unsupervised CLIR approach; see \S\ref{sec:base_results_docs}) and (b) after in-domain fine-tuning on English CLEF data via contrastive metric-based learning (see \S\ref{sec:l2r}): here we have only zero-shot language transfer, but no domain transfer (as was the case with L2R models from the previous section).}
    \label{fig:contrastive}
\end{figure}

We now empirically investigate the second common scenario in ad-hoc retrieval: a limited amount of ``in-domain'' relevance judgments that can be leveraged for fine-tuning of text encoders (as opposed to a large amount of ``out-of-domain'' training data sufficient to train full-blown learning-to-rank classifiers, covered in the previous subsection). To this end, we use the relevance judgments in the English portion of the CLEF collection to fine-tune our best-performing multilingual encoder ($\text{DISTIL}_{\text{DmBERT}}$), using the contrastive metric-based learning objective (see \S\ref{sec:l2r}) to refine the representation space of the encoder. 
We carry out fine-tuning and evaluation in a 10-fold cross-validation setup (i.e., we carry out fine-tuning 10 different times, each time training on different nine-tenths of the queries and evaluating on the remaining one-tenth) in order to prevent any information leakage between languages: in the CLEF collection, queries in languages other than English are simply translations of the English queries. This resulted (in each fold) with a fine-tuning training set consisting of merely $800$-$900$ positive instances (in English). We trained in batches of $16$ positive instances and for each of them created all possible in-batch negatives\footnote{This means at most $15$ in-batch negatives created from the other query-document pairs in the batch; there is less than $15$ negatives only if there are other positive instances for the same query in the batch.} for the Multiple Negative Ranking Loss objective (see \S\ref{sec:l2r}).   
With cross-validation in place, for each language pair, we obtain predictions for all queries without any information leakage, which makes the results of contrastive fine-tuning fully comparable with all previous results. 

The CLIR results of the ranking with the contrastively fine-tuned $\text{DISTIL}_{\text{DmBERT}}$ are shown in Figure \ref{fig:contrastive}. Unlike re-ranking with full-blown pointwise learning to rank models from the previous section, contrastive in-domain reshaping of the representation space of the multilingual encoder yields performance gains for all language pairs (2.5 MAP points on average). It is important to emphasize again that -- because contrastive metric-based fine-tuning only updates the parameters of the original multilingual transformer ($\text{DISTIL}_{\text{DmBERT}}$) and introduces no additional parameters (i.e., no classification head on top of the encoder, as in the case of L2R models trained on MS MARCO and TREC ROBUST from the previous section) -- we can, in exactly the same manner as with the base model before fine-tuning, fully rank the entire document collection for a given query, instead of restricting ourselves to re-ranking the top results of the base ranker. 

Summarizing the results from this section and the previous one, it appears that -- at least when it comes to zero-shot language transfer for cross-lingual document retrieval -- specializing the representation space of a multilingual encoder with few(er) in-domain relevance judgments is more effective than employing a neural L2R ranker trained on large amounts of ``out-of-domain'' data.

\subsection{Cross-Lingual Retrieval or Cross-Lingual Transfer for Monolingual Retrieval?}
\label{sec:multi_vs_cross}

At first glance, our negative CLIR results for the mBERT-based pointwise L2R rankers (\S\ref{sec:reranking}) -- i.e., the fact that using them for re-ranking does not improve the performance of our best-performing unsupervised ranker ($\text{DISTIL}_{\text{DmBERT}}$) -- seem at odds with their solid cross-lingual transfer results reported in previous work \cite{macavaney2020teaching}. It is, however, important to notice the fundamental difference between two evaluation settings: what was previously evaluated \cite{macavaney2020teaching} was the effectiveness of (zero-shot) \textit{cross-lingual transfer} of a \textit{monolingual retrieval} model, trained on English data and transferred to a set of target languages. In other words, both in training and at inference time the models deal with queries and documents written in the same language. Our work here, instead, focuses on a fundamentally different scenario of cross-lingual retrieval, where the language of the query is different from the language of document collection. 
We argue that, in a supervised setting, in which one trains on monolingual English data only, the latter (i.e., CLIR) represents a more difficult transfer setup. 

To validate the above assumption, we additionally evaluate the two mBERT-based re-rankers from \S\ref{sec:reranking} trained on MS MARCO and TREC ROBUST, respectively, on monolingual portions of the CLEF collection. We use them to re-rank two strong  monolingual baselines: (1) Query Likelihood Model (QLM, based on unigrams) \cite{ponte1998language} with Dirichlet smoothing \cite{zhai2004study}, which we also used for the machine-translation baseline (MT-IR) in our base evaluation (see \S\ref{sec:exp_setup_unsup}); and (2) a retrieval model based on aggregation of IDF-scaled static word embeddings (\S\ref{sec:clwes}; Eq.~\eqref{eq:static_wemb_eq}).\footnote{This corresponds to the Proc-B baseline in CLIR evaluations; only here we use monolingual embeddings of the target language (instead of a bilingual word embedding space, as in CLIR).} For the latter, we used the monolingual FastText embeddings trained on Wikipedias of respective languages,\footnote{\url{https://fasttext.cc/docs/en/pretrained-vectors.html}} with vocabularies limited to the 200K most frequent terms.       

\begin{table*}[t!]
\centering
\caption{Cross-lingual zero-shot transfer for monolingual retrieval: results on the monolingual CLEF portions. Base rankers (top third of the table) -- QLM with Dirichlet Smoothing and aggregation of static monolingual word embeddings (FastText) and re-ranking with pointwise mBERT-based models trained on English MS MARCO (middle third) and TREC ROBUST data (bottom third), respectively.} 
\vspace{1mm}
\def\arraystretch{0.95}
{\scriptsize
{
\begin{tabularx}{\linewidth}{l X X X X X X X} 
\toprule
 & EN-EN & FI-FI & DE-DE & IT-IT & RU-RU & AVG & $\Delta$ AVG \\ \midrule
\textit{No re-ranking (reference)} \\ \midrule 
QLM & .471 & .376 & .400 & .463 & .325 & .407 & -- \\
FastText & .310 & .327 & .314 & .314 & .214 & .296 & -- \\ \midrule
\textit{Re-ranker trained on MS MARCO} \\ \midrule 
QLM & .433 & .349 & .367 & .382 & .290 & .364 & $-$4.27 \\ 
FastText & .371 & .341 & .306 & .333 & .296 & .329 & $+$3.36 \\ \midrule
\textit{Re-ranker trained on TREC ROBUST} \\ \midrule 
QLM & .481 & .520 & .420 & .454 & .303 & .436 & $+$1.98 \\
FastText & .375 & .462 & .367 & .429 & .299 & .386 & $+$8.76 \\ \bottomrule
\end{tabularx}
}}
\label{tab:monolingual_transfer_rerank}
\vspace{-1.5mm}
\end{table*}

The results of mBERT-based re-rankers in cross-lingual transfer for monolingual retrieval are summarized in Table \ref{tab:monolingual_transfer_rerank}. We see that, unlike in CLIR (see Table \ref{tab:clefrerank}), mBERT-based re-rankers do substantially improve the performance of the base retrieval models (the only exception is re-ranking of QLM results with the MS MARCO model), even despite the fact that the base performance of the monolingual baselines (QLM and FastText) is significantly above the best CLIR performance we observed with unsupervised rankers (see $\text{DISTIL}_{\text{DmBERT}}$ in Table \ref{tab:clefselfsup}). This is in line with the findings from \cite{macavaney2020teaching}: multilingual encoders (e.g., mBERT) do seem to be a viable solution for (zero-shot) cross-lingual transfer of learning-to-rank models for monolingual retrieval. 
But why are they not as effective when transferred to CLIR settings (as shown in \ref{sec:reranking})? We hypothesize that monolingual English training on large-scale datasets like MS MARCO or TREC ROBUST leads to a sort of ``overfitting'' to monolingual retrieval (e.g., the model may implicitly learn to assign a lot of importance to exact term matches) -- such (latent) features will, in principle, transfer reasonably well to other monolingual retrieval settings, regardless of the target language; with queries in different language from documents, however, CLIR instances are likely to generate out-of-training-distribution values for these latent features (e.g., if the model learned to value exact matches during training, at predict time in CLIR settings, it would need to recognize word-level translations between the two languages), confusing the pointwise classifier.         



%% file: 6-conclusion.tex
\section{Conclusion}


Pretrained multilingual encoders have been shown to be widely useful in natural language understanding (NLU) tasks, when fine-tuned in supervised settings on some task-specific data; their utility as general-purpose text encoders in unsupervised settings, such as the ad-hoc cross-lingual IR, has been less investigated. In this work, we systematically validated the suitability of a wide spectrum of cutting-edge multilingual encoders for document- and sentence-level CLIR across diverse languages. 

We first profiled the popular self-supervised multilingual encoders (mBERT and XLM) as well as the multilingual encoders specialized for semantic text matching on semantic similarity datasets and parallel data as text encoders for unsupervised CLIR. Our empirical results show that self-supervised multilingual encoders (mBERT and XLM), without exposure to task supervision, generally fail to outperform CLIR models based on static cross-lingual word embeddings (CLWEs). Semantically-specialized multilingual sentence encoders, on the other hand, do outperform CLWEs; the gains, however, are pronounced only in sentence retrieval, while being much more modest in document retrieval. 

Acknowledging that sentence-specialized multilingual encoders are not designed for encoding long documents, we proposed to exploit their strength -- precise semantic encoding of short texts -- in document retrieval too, by means of localized relevance matching, where we compare the query with individual document segments or sentences and max-pool the relevance scores; we showed that such localized relevance matching with sentence-specialized multilingual encoders yields substantial document-level CLIR gains. 

Finally, we investigated how successful supervised (re-)rankers based on multilingual encoders are in ad-hoc CLIR evaluation settings. We show that, while rankers trained monolingually on large-scale English datasets (e.g., MS-MARCO) can be successfully transferred to monolingual retrieval tasks in other languages by means of multilingual encoders, their transfer to CLIR setups, in which the query language differs from the language of the document collection, is much less successful. Furthermore, we introduced an alternative supervised approach, based on contrastive metric-based learning, designed for fine-tuning the representation space of a multilingual encoder when only a limited amount of ``in-domain'' relevance judgments is available. We show that such small-scale in-domain fine-tuning of multilingual encoders yields better CLIR performance than rankers trained on large external collections (i.e., out-of-domain). 

While state-of-the-art multilingual text encoders excel in so many seemingly more complex language understanding tasks, our work renders ad-hoc CLIR in general and document-level CLIR in particular a serious challenge for these models. \added{Simple bag-of-words retrieval with the Query Likelihood Model on the machine translated query (MT-IR) remains a strong baseline outperforming all semantic encoding variations explored in document retrieval.} We believe that our systematic comparative evaluation of a multitude of multilingual encoders (as both unsupervised and supervised rankers) offers a multitude of insights for practitioners dealing with (ad-hoc) cross-lingual retrieval task. While there are scenarios in which multilingual encoders can substantially improve CLIR performance, our work identifies potential pitfalls and emphasizes conditions needed for solid CLIR performance with multilingual text encoders. We make our code and resources available at \url{https://github.com/rlitschk/EncoderCLIR}.